\documentclass{article} 
\usepackage{iclr2023_conference,times}

\usepackage{titletoc}
\newcommand\DoToC{%
  \startcontents
  \printcontents{}{1}{\hrulefill\vskip0pt}
  \vskip0pt \noindent\hrulefill
  }

\usepackage{amsmath,amsfonts,bm}

\def\eqref#1{equation~\ref{#1}}

\def\1{\bm{1}}

\DeclareMathAlphabet{\mathsfit}{\encodingdefault}{\sfdefault}{m}{sl}
\SetMathAlphabet{\mathsfit}{bold}{\encodingdefault}{\sfdefault}{bx}{n}

\newcommand{\R}{\mathbb{R}}

\usepackage{url}
\usepackage[utf8]{inputenc} 
\usepackage[T1]{fontenc}    
\definecolor{mydarkblue}{rgb}{0,0.08,0.45}
\usepackage{hyperref}       
\hypersetup{ %
    pdftitle={},
    pdfauthor={},
    pdfsubject={},
    pdfkeywords={},
    pdfborder=0 0 0,
    pdfpagemode=UseNone,
    colorlinks=true,
    linkcolor=black,
    citecolor=mydarkblue,
    filecolor=mydarkblue,
    urlcolor=mydarkblue,
    pdfview=FitH}
\usepackage{url}            
\usepackage{booktabs}       
\usepackage{amsfonts}       
\usepackage{nicefrac}       
\usepackage{microtype}      
\usepackage{xcolor}         

\usepackage{multirow}
\usepackage{subcaption}
\usepackage{bigstrut}
\usepackage{graphicx}
\usepackage{amsmath}
\usepackage{amssymb}
\usepackage{algorithm}
\usepackage{algorithmicx}
\usepackage{algpseudocode}
\usepackage{tabu}
\usepackage{siunitx}
\usepackage{adjustbox}
\usepackage{subcaption}
\usepackage{siunitx}
\usepackage{makecell}
\usepackage{xcolor}
\usepackage{colortbl}
\usepackage{color}
\usepackage{multirow}
\usepackage{blindtext}
\usepackage{comment}
\usepackage{bm}
\usepackage{bbm}
\usepackage{booktabs}
\usepackage{wrapfig}
\usepackage{array} 
\definecolor{Gray}{gray}{0.95}
\definecolor{Cyan}{rgb}{0.88,1,1}
\definecolor{LightCyan}{rgb}{0.92,1,1}
\definecolor{DarkCyan}{rgb}{0.82,1,1}
\newcommand{\ourmeos}{\textbf{\texttt{PLOT}} }

\usepackage{wrapfig}

\newcommand{\tabstyle}[1]{
  \setlength{\tabcolsep}{#1}
  \centering
  \small
}
\newcommand{\ourtitle}{PLOT: Prompt Learning with Optimal Transport for Vision-Language Models}
\title{\ourtitle}
\usepackage{appendix}

\author{Guangyi Chen$^{\dagger \bullet}$, Weiran Yao$^{\dagger}$, Xiangchen Song$^{\dagger}$, Xinyue Li$^{\diamond}$, Yongming Rao$^{\ddagger}$, Kun Zhang$^{\dagger \bullet}$ \\
$^{\dagger}$Carnegie Mellon University, Pittsburgh PA, USA \\
$^{\bullet}$Mohamed bin Zayed University of Artificial Intelligence, Abu Dhabi, UAE \\
$^{\ddagger}$Tsinghua University, Beijing, China \\
$^{\diamond}$New York University, Abu Dhabi, UAE \\}

\iclrfinalcopy 
\begin{document}

\maketitle

\begin{abstract}
  With the increasing attention to large vision-language models such as CLIP, there has been a significant amount of effort dedicated to building efficient prompts. Unlike conventional methods of only learning one single prompt, we propose to learn multiple comprehensive prompts to describe diverse characteristics of categories such as intrinsic attributes or extrinsic contexts. However, directly matching each prompt to the same visual feature is problematic, as it pushes the prompts to converge to one point. To solve this problem, we propose to apply optimal transport to match the vision and text modalities. Specifically, we first model images and the categories with visual and textual feature sets. Then, we apply a two-stage optimization strategy to learn the prompts. In the inner loop, we optimize the optimal transport distance to align visual features and prompts by the Sinkhorn algorithm, while in the outer loop, we learn the prompts by this distance from the supervised data. Extensive experiments are conducted on the few-shot recognition task and the improvement demonstrates the superiority of our method. The code is available at \url{https://github.com/CHENGY12/PLOT}.
\end{abstract}

\vspace{-10pt}
\section{Introduction}
\vspace{-10pt}
In the past few years, large-scale vision-language pre-trained (VLP) models, such as CLIP~\citep{clip}, ALIGN~\citep{align}, and BLIP~\citep{blip} have achieved remarkable success in open-world visual concept learning. These methods have brought new light but also pose a new question: how to efficiently adapt the knowledge from pretraining to the downstream tasks since these models are typical of massive sizes which are not feasible for normal users to re-train.


\begin{wrapfigure}{r}{9cm}
\vspace{-10px}
    \includegraphics[width=9cm]{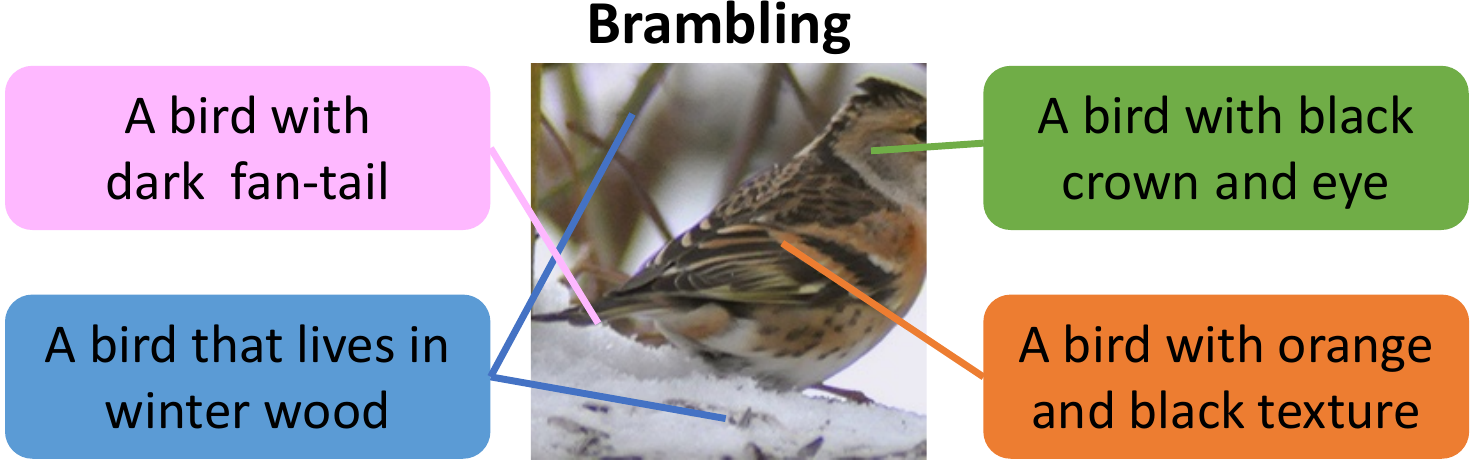}
    \caption{The motivation that one category can be complementarily described in different views (An example of ``Brambling'').   }
    \label{fig:motivation}
\vspace{-10px}
\end{wrapfigure}
One of the conventional paradigms of utilizing pretrained knowledge is ``pre-training, fine-tuning'', which fixes the architecture of the pre-trained neural network and tunes its parameters using task-specific objective functions. Beyond fine-tuning the parameters, many recent methods~\citep{coop,cocoop} introduce the concept of prompt learning from the field of NLP to the vision domain and achieve striking performance gain for the few-shot visual classification. They fix the model parameters and instead learn suitable prompts by turning a template sentence into a set of learnable vectors. Then, these prompts are learned by minimizing the distance between the visual features and prompt-based language features.

Despite significant improvements over manual prompts, learning only a sentence is intuitively insufficient to represent a class.  One class can be described by many intrinsic characteristics and even extrinsic context relations. Thus, for one object, we may have multiple prompt candidates which focus on different attributes. As shown in Figure~\ref{fig:motivation}, we can describe the class ``Brambling'' in different views: such as the color of the wing, the color of the crown and eyes, the shape and color of the tail, and even the living environment information. 
It motivates us to learn multiple prompts to comprehensively represent the class and thus facilitate classification.

The most natural solution is to directly learn multiple prompts by respectively matching each prompt with the visual features. However, it is the same as matching the mean of prompt features and the visual features. This solution is problematic since all prompts are encouraged to be closer to one single point and thus tend to learn the same characteristics. It contradicts our purpose to learn comprehensive prompts. To solve this problem, we tested adding some constraints to push away the prompt from each other, but found that this solution still fails to learn representative and comprehensive prompts. This solution treats the visual representation as one single point, and such a unified view of visual features ignores the fact that different prompts may only focus on one or a subset of characteristics.


To address this problem, in this paper, we propose Prompt Learning with Optimal Transport (\texttt{\textbf{PLOT}}), which applies optimal transport~(OT) to align the local visual features and multiple textual prompts. Optimal transport can calculate the distance between two distributions under the form of multiple sampling. In our prompt learning framework, we formulate local visual features and multiple prompts as the samplings of two discrete distributions and use OT to encourage fine-grained cross-modal matching.
Specifically, to obtain the local visual features with different semantic clues, we extract all feature maps as the visual representation instead of the single global representation. 
Fortunately, we can easily obtain the visual feature maps from the visual encoder of CLIP by using all outputs of the multi-head self-attention layer~\citep{rao2021denseclip}.
Then the problem comes down to how to calculate the distance between two feature sets. 

We solve this problem by introducing the optimal transport theory~\citep{villani2009optimal} and formulate the feature sets as a discrete probability distribution where each feature has an equal probability value. 
Furthermore, to reduce the computational cost and avoid the extra model parameters, we learn the prompts with a two-stage optimization strategy. 
At the first stage in the inner loop, we fix both visual and text features and optimize the optimal transport problem by a fast Sinkhorn distances algorithm~\citep{cuturi2013sinkhorn}. Then, in the outer loop, we fix all parameters of optimal transport and back-propagate the gradient to learn the prompts with different characteristics. Compared with conventional distance (such as Euclidean distance of mean features), optimal transport can align different visual features for each local prompt, which is more robust to the visual misalignment and tolerates well feature shift~\citep{emd}. It is because OT learns an adaptive transport plan to align features, which achieves fine-grained matching across two modalities. We conduct experiments on 11 datasets following the standard setting of CLIP~\citep{clip} and CoOp~\citep{coop} to evaluate our method. These experiments span the visual classification of generic objects, scenes, actions, fine-grained categories, and so on. The significant result improvement demonstrates that \ourmeos can effectively learn representative and comprehensive prompts. 


\vspace{-10pt}
\section{Related Work}
\vspace{-5pt}
\textbf{Optimal Transport}
The Optimal Transport~\citep{monge1781memoire} is initially introduced to solve the problem of how to reduce the cost when moving several items simultaneously. 
Recently, OT theory has drawn wide attention in the machine learning and computer vision community by comparing distributions readily available to them under the form of feature sets~\citep{COTFNT}. Due to the brilliant property of distribution matching, OT has been applied in many theoretic and application tasks including 
generative models~\citep{arjovsky2017wasserstein,salimans2018improving,zhao2020neural}, structural matching~\citep{chen2019improving,xu2020vocabulary,zhao2021towards,xu2019gromov} (e.g. sequence matching~\citep{chen2019improving} and graph matching~\citep{xu2019gromov}, and image matching~\citep{zhang2020deepemd,liu2021multi,zhao2021towards}), and other distribution-based tasks (such as clustering~\citep{laclau2017co}, distribution estimation~\citep{boissard2015distribution}, and causal discovery~\citep{tu2022optimal}). In this paper, we use OT to align the features of vision and language modalities 
by learning an adaptive transport plan~\citep{emd}.

\textbf{Vision-Language Pre-trained Models}
Vision-Language Pre-trained (VLP) models aim to explore the semantic correspondence between the vision and language modalities through large-scale pre-training. Recently, VLP models have achieved an exciting performance improvement in few-shot visual recognition~\citep{clip,clip-adapter,coop,cocoop,vtclip}, which shows the great potential to promote open-world visual understanding with the help of language. 
In terms of objectives, VLP methods can be divided into reconstruction~\citep{visualbert,vln-bert,meter,vilt}, contrastive matching~\citep{clip,align,mural}, or the combination of both two~\citep{li2021align,vlmo,kamath2021mdetr}. Besides, recent progress in the field of VLP also benefits a lot from large-scale pair-wised datasets. For example, CLIP~\citep{clip} applies 400 million image-text pairs for contrastive learning. Beyond recognition, these VLP models also show great potential for other downstream applications, such as dense prediction~\citep{rao2021denseclip,zhou2021denseclip}, image generation~\citep{nichol2021glide,ramesh2022hierarchical,patashnik2021styleclip}, and action understanding~\citep{wang2021actionclip,tevet2022motionclip}.

\textbf{Prompt Learning}
Prompt learning is introduced from the field of NLP to efficiently adapt the large language model to downstream tasks. Different from the conventional ``pre-training, fine-tuning'' paradigm which initializes the pre-trained model and tunes the parameters of the network using downstream task-specific objective functions, prompt learning applies textual prompt to reformulate the downstream tasks as the original pretrained task~\citep{liu2021pre,petroni2019language}. 
By the prompt, the domain shift between pretrained task and the downstream application is reduced and thus the pretrained knowledge can be easier adapted to downstream tasks. 
The concept of prompt learning~\citep{petroni2019language,radford2019language,poerner2019bert} begins from the success of GPT~\citep{radford2019language} series. Early prompt learning methods (such as Petroni~\emph{et al.}~\citep{petroni2019language} and Pörner~\emph{et al.}~\citep{poerner2019bert}) always manually create templates based on human prior knowledge. 
Furthermore, some mining-based methods~\citep{jiang2020can} and gradient-based methods~\citep{autoprompt} are proposed to automatically search for appropriate templates. Beyond search in the discrete space, some methods~\citep{li2021prefix,tsimpoukelli2021multimodal,liu2021gpt} remove the constraint that the prompts are ``words'' and instead learn prompts in the continuous embedding space. Recently, CoOp~\citep{coop} and its extended version~\citep{cocoop} introduce prompt learning into open-world visual understanding to adapt the knowledge from the large-scale visual-language pretrained models and achieve great performance improvement on the few-shot visual recognition.
Compared with CoOp, our \ourmeos method further improves prompt learning by introducing the optimal transport distance to learn multiple local prompts and achieves fine-grained vision-language matching.
PDL~\cite{pdl} is also motivated by the more diverse prompts, which assumes a parametric distribution of prompts and fits the parameters during training. Different from it, \ourmeos learns multiple prompts without parametric distribution.

\vspace{-10pt}
\section{Approach}
\vspace{-5pt}
In this section we first revisit the baseline CoOp~(\ref{sec:revisit CoOp}), review the preliminaries of optimal transport~(\ref{sec:Optimal Transport}), and then introduce our \ourmeos (\ref{sec:PLOT}) to show how we learn multiple comprehensive prompts. 
\vspace{-10pt}
\subsection{A Revisit of CoOp} \label{sec:revisit CoOp}
\vspace{-5pt}
%

CoOp~\citep{coop} is one of the pioneering methods to learn the prompts for using vision language pretrained knowledge (such as CLIP~\citep{clip}) for downstream open-world visual recognition. Different from CLIP which manually designs the prompt templates, CoOp sets a part of context words in the template as continuous learnable parameters which can be learned from the few-shot data.
Then the classification weights can be represented by the distance between the learned prompt and visual feature. 

Specifically, given an image $\bm{x}$, a visual feature $\bm{f}=f(\bm{x})$ is obtained by
the visual encoder $f$ of CLIP. 
Then, the textual prompt can be formulated as $\bm{t}_k = \{\bm{\omega}_1,\bm{\omega}_2,\hdots,\bm{\omega}_L, \bm{c}_k \} $, where $\bm{c}_k$ is the word embedding of the class name, $ \bm{\omega} =\{\bm{\omega}_l|_{l=1}^{L}\}$ are learnable vectors where each vector has the same dimension as the original word embedding and L is the length of context words.
With prompt $\bm{t}_k $ as the input, the text encoder $g$ outputs the textual feature as $\bm{g}_k=g(\bm{t}_k)$. 
The final prediction probability is computed by the matching score as follows:
\begin{equation} \label{eq:predict}
p(y=k|\bm{x}) = \frac{\exp(\text{sim}(\bm{f},\bm{g}_k)/\tau)}{\sum_{k'=1}^{K}\exp(\text{sim}(\bm{f},\bm{g}_{k'})/\tau) },
\end{equation}
where $\text{sim}(\cdot,\cdot)$ denotes a metric function such as cosine similarity, and $\tau$ stands for the temperature of Softmax.
Then we can optimize the parameters of $\{\bm{vec}_l|_{l=1}^{L}\}$ with the cross-entropy loss between the prediction and the labeled target. 

\begin{figure}[t]
    \centering
    \includegraphics[width=\textwidth]{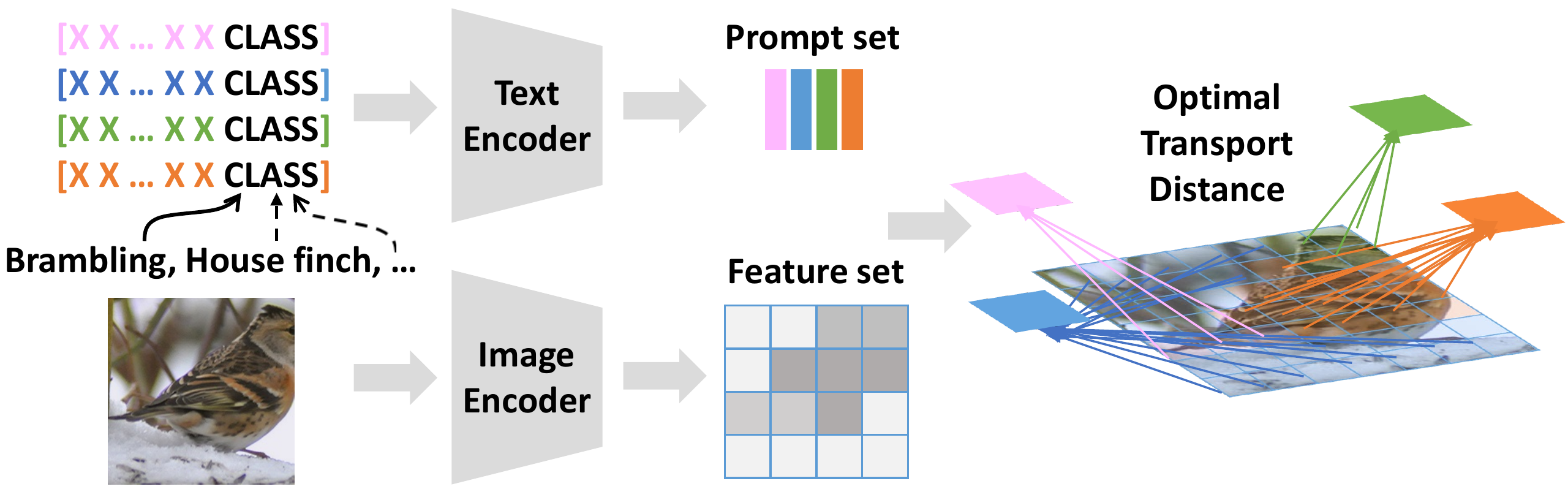}
    \caption{\textbf{The framework:} \ourmeos first describes each category with multiple prompts and obtains a set of prompt features by text encoder. The image is also encoded as a set of local features. Then the optimal transport is used as the metric between prompts and visual features.}
    \label{fig:plot}
    \vspace{-10px}
\end{figure}
\vspace{-5pt}
\subsection{Optimal Transport} \label{sec:Optimal Transport}
\vspace{-5pt}
Optimal transport~(OT) distance is a widely used metric for the comparison of distributions. Here, we only focus on the discrete situation which is more related to our framework. Assuming we have two sets of points (features), the discrete distributions are formulated as:
\begin{equation} \label{eq:discrete}
U=\sum_{m=1}^{M}u_m\delta_{\bm{f}_m} \hspace{2em} \text{and} \hspace{2em} V=\sum_{n=1}^{N}v_n\delta_{\bm{g}_n},
\end{equation}
where $\bm{u}$ and $\bm{v}$ are the discrete probability vectors that sum to 1, and $\delta_{\bm{f}}$ is a Dirac delta function placed at support point $\bm{f}$ in the embedding space.
Then, the total distance is written as:
\begin{equation} \label{eq:cost}
<\bm{T},\bm{C}> = \sum_{m=1}^{M}\sum_{n=1}^{N}\bm{T}_{m,n}\bm{C}_{m,n}.
\end{equation}
We call $\bm{C}$ the cost matrix in which each point denotes the cost between $\bm{f}_m$ and $\bm{g}_n$, such as $\bm{C}_{m,n}= 1- \text{sim}(\bm{f}_m,\bm{g}_{n})$. While the $\bm{T} $ is called the transport plan, which is learned to minimize the total distance.
The optimization problem of optimal transport is formulated as:
\begin{equation} \label{eq:optimization}
\begin{aligned}
& d_{\text{OT}}(\bm{u},\bm{v}|\bm{C}) = \underset{\bm{T}}{\text{minimize}}
 <\bm{T},\bm{C}> \\
& \text{subject to}
~~~~~~\bm{T}\bm{1}_N = \bm{u},\; \bm{T}^\top \bm{1}_M = \bm{v},\; \bm{T} \in \R^{M \times N}_+.
\end{aligned}
\end{equation}

As directly optimizing the above objective is always time-consuming, we apply the Sinkhorn distance~\citep{cuturi2013sinkhorn} to use an entropic constraint for fast optimization.
The optimization problem with a Lagrange multiplier of the entropy
constraint is:
\begin{equation} \label{eq:Sinkhorn}
\begin{aligned}
& d_{\text{OT},\lambda}(\bm{u},\bm{v}|\bm{C})=\underset{\bm{T}}{\text{minimize}}
 <\bm{T},\bm{C}> - \lambda h(\bm{T})\\
& \text{subject to}
~~~~~~~\bm{T}\bm{1}_N = \bm{u},\; \bm{T}^\top \bm{1}_M = \bm{v},\;
\bm{T} \in \R^{M \times N}_+,
\end{aligned}
\end{equation}
where $h(\cdot) $ is entropy and $\lambda \geq 0$ is a hyper-parameter. Then we can have a fast optimization solution with a few iterations as:
\begin{equation} \label{eq:Sinkhorn_optimization}
\bm{T}^*= \text{diag}(\bm{u}^{(t)}) \exp(-\bm{C}/\lambda) \text{diag}(\bm{v}^{(t)}),
\end{equation}
where $t$ denotes the iteration and in each iteration 
$\bm{u}^{(t)} =\bm{u}/\left((\exp(-\bm{C}/\lambda)\bm{v}^{(t-1)}\right)$ and  $\bm{v}^{(t)} =\bm{v}/\left((\exp(-\bm{C}/\lambda)^\top\bm{u}^{(t)}\right) $, with the initiation $\bm{v}^{(0)} = \bm{1}$.

\vspace{-5pt}
\subsection{Prompt Learning with Optimal Transport}
\label{sec:PLOT}
\vspace{-5pt}
In this subsection, we introduce the details of our \ourmeos, which learns multiple prompts to describe different characteristics of the category by minimizing the OT distance.

Specifically, as shown in Figure~\ref{fig:plot}, given an image $\bm{x}$, we first feed it to the visual encoder branch of CLIP. Apart from the global visual feature $\bm{f}$, we can also obtain a set of local features $\{\bm{f}_m|_{m=1}^{M}\}$. The visual encoder has a multi-head attention pooling layer in which the input is the combination of the global feature and a set of local features (feature map) and the output is a tensor with the shape $\mathbb{R}^{(H\times W+1)\times C}$, where $H$ and $W$ is the height and width of feature map and $C$ is the feature dimension. 
Therefore, we can obtain $M=H\times W$ local features and a global feature. At the same time, for class $k$, we can initialize N local prompts as $\{\bm{t}_{k,n}|_{n=1}^{N}\} $ with learnable vectors $\{\bm{\omega}_{n}|_{n=1}^{N}\}$, where each is the same as the prompt in CoOp. With both visual and textual encoders, we can obtain local visual features $\bm{F}=\{\bm{f}_m|_{m=1}^{M}\} \in \mathbb{R}^{M\times C}$ and prompt features $\bm{G}_k=\{\bm{g}_n|_{n=1}^{N}\} \in \mathbb{R}^{N\times C}$. 

In the inner loop, we learn the transport plan $\bm{T}$ with these fixed support sets $\bm{F},\bm{G}_k$, by minimizing the following OT distance to push $\bm{G}_k$ to $\bm{F}$:
\begin{equation} \label{eq:distance}
\begin{aligned}
 d_{\text{OT}}(k) = d_{\text{OT}}(\bm{u},\bm{v}|\bm{1}-\bm{F}^\top\bm{G}_k),
\end{aligned}
\end{equation}
where $\bm{C}=\bm{1}-\bm{F}^\top\bm{G}_k$ denotes that we use the cosine distance between $\bm{F}$ and $\bm{G}_k$ as the cost matrix. Then we can obtain the solution of transport plan $\bm{T}^*$ as Eq.~\ref{eq:Sinkhorn_optimization} and the final OT distance $d_{\text{OT}}(k)$.

Given the OT distance between $\bm{G}_k$ and $\bm{F}$, we reformulate the prediction probability as:
\begin{equation} \label{eq:predict_OT}
p_{\text{OT}}(y=k|\bm{x}) = \frac{\exp \left(\left(1-d_{\text{OT}}(k)\right)/\tau \right)}{\sum_{k'=1}^{K} \exp\left(\left(1-d_{\text{OT}}(k')\right)/\tau\right) }.
\end{equation}
In the outer loop, we fix the transport plan $\bm{T}^*$ and optimize $\{\bm{vec}_{l,n}|_{l=1,n=1}^{L,N}\}$ with cross entropy:
\begin{equation} \label{eq:loss}
L_{\text{CE}} =-\frac{1}{|\mathcal{X}|}\sum_{\bm{x} \in \mathcal{X}} \sum_{k=1}^{K} y_{\bm{x},k} p_{\text{OT}}(y=k|\bm{x}),
\end{equation}
where $\bm{y}_{\bm{x}}$ is a one-hot label vector. The detailed algorithm can be found in Appendix~\ref{app:ot}. 

Though the optimization strategy of the optimal transport and prompts is two-stage, the whole training flow is end-to-end. It is because the transport plan is computed using a small number of matrix multiplications as a forward module. The gradients of these matrix multiplications are taped for back-propagation for end-to-end optimization, which makes the whole system fully differentiable (including the iterative algorithm) and easy to implement using an autograd library like PyTorch.
In the experiments, we found that it is natural and relatively easy to this optimization strategy.

 

\vspace{-10pt}
\section{Experiments}
\vspace{-5pt}
Extensive experiments are conducted to evaluate our method, including comparison with CoOp, ablation studies, parameter analysis extensibility analysis, computing cost analysis, and visualization. 
\vspace{-10pt}
\subsection{Datasets \label{sec:datasets}}
\vspace{-5pt}
We followed the experimental settings in the CoOp~\citep{coop} for the few-shot learning evaluation. The experiments are conducted on the 11 visual recognition datasets, including Caltech101~\citep{fei2004learning}, DTD~\citep{cimpoi2014describing}, EuroSAT~\citep{helber2019eurosat}, FGVCAircraft~\citep{maji2013fine}, Flowers102~\citep{nilsback2008automated}, Food101~\citep{bossard2014food}, ImageNet~\citep{deng2009imagenet}, OxfordPets~\citep{parkhi2012cats}, StanfordCars~\citep{cars}, SUN397~\citep{xiao2010sun}, and UCF101~\citep{soomro2012ucf101}. These datasets span visual classification of generic objects, scenes, actions, fine-grained categories, and so on, which constitutes a comprehensive evaluation of our method. All experiments adopted the few-shot evaluation protocol used in CLIP~\citep{clip} and CoOp~\citep{coop}, where we respectively choose 1, 2, 4, 8, and 16 shots for model training and use the original test set for evaluation. Besides, we also evaluated the robustness of our method with domain shift. Following CoOp, we used the ImageNet as the source domain and evaluate our method with ImageNet-based robustness evaluation datasets including ImageNetV2~\citep{ImageNetv2}, ImageNet-Sketch~\citep{wang2019learning}, ImageNet-A~\citep{imagenet-a}, and ImageNet-R~\citep{imagenet-r}. 
A detailed introduction of each dataset can be found in the appendix. 

\vspace{-5pt}
\subsection{Implementation details}
\label{sec:details}
\vspace{-5pt}

We chose CoOp~\citep{coop} as our main competitor to evaluate our method. Compared with CoOp which only learns a global prompt for one class, our \ourmeos method learns multiple local prompts and applies the OT distance for fine-grained alignment.
Besides, we also reported the performance of training a linear classifier with the CLIP~\citep{clip} features and the conditional version of CoOp, called CoCoOp~\citep{cocoop}. They are also widely-used methods to adapt the pretrained knowledge for the downstream task. Please note that we evaluate CoCoOp in the same setting for a fair comparison (the base-to-new setting can be found in the appendix).  
The original CoOp method has different versions with different class token positions and parameter initialization strategies.  For easy comparison, we directly chose one of them as our baseline with ``end'' token position, ``random'' initialization, 16 context tokens, and RN50 backbone. More implementation details can be found in Section~\ref{app:implementation}.

\begin{figure}
  \centering
  \includegraphics[width=0.92\linewidth]{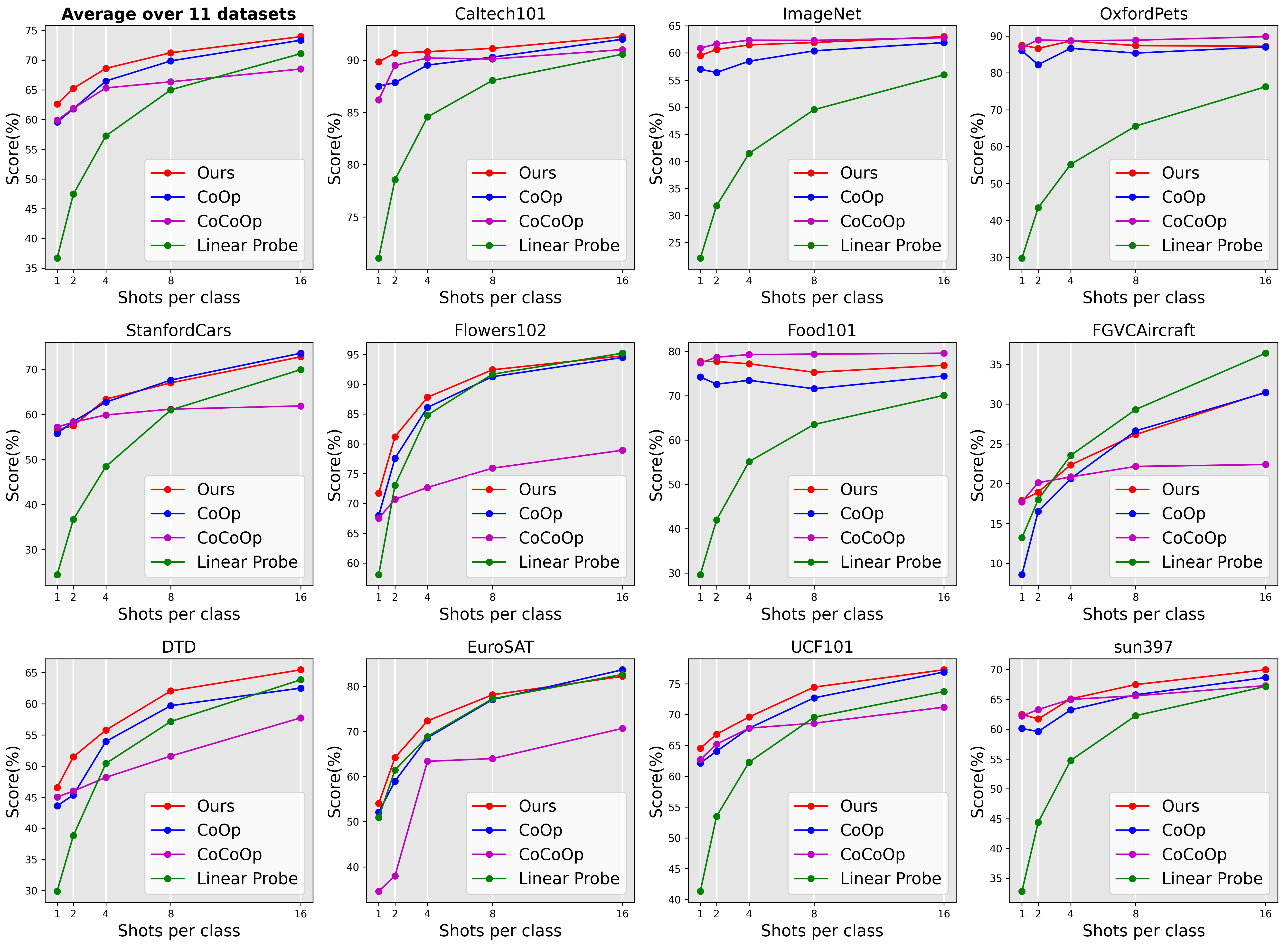}
  \caption{ The few-shot learning results on 11 datasets. We compare our \ourmeos with CoOp, CoCoOp, and the Linear Probe method and observe the consistent and significant performance improvement on most datasets. (The average accuracy on all datasets is shown on the left top.)}
\label{fig: comparison_curve}
\vspace{-0.5cm}
\end{figure}
\vspace{-5pt}
\subsection{Comparison With CoOp}
\vspace{-5pt}
In this subsection, we compare our \ourmeos with the baseline CoOp on the few-shot recognition and domain generalization tasks.

\textbf{Few-Shot Learning}
We summarized the experimental results in Figure~\ref{fig: comparison_curve} where the red line denotes our \ourmeos method, the blue one denotes CoOp, the purple line denotes CoCoOp~\citep{cocoop}, and the green one is the CLIP linear probe. As the settings in the CoCoOp and CoOp are different, we re-run the CoCoOp method in the setting of CoOp.
We observed that all prompt learning methods outperform the linear probe method by a large margin. Besides, \ourmeos can further outperform CoOp and CoCoOp on most of the datasets. 
Taking the average accuracy (at the left top) as the example, \ourmeos respectively gained $3.03\%, 3.45\%, 2.13\%, 1.38\%, 0.61\%$ performance boost over CoOp at $1,2,4,8,16$ shots.
Among all datasets, \ourmeos achieves a larger improvement over CoOp on the FOOD101 and DTD datasets and achieves comparable performance on the StanfordCars datasets. 
It may be because the discriminative characters in StanfordCars coincide with each other, such that one global prompt and one global visual feature can work well.
Note that we don't use the class-specific context, thus the performance on the fine-grained classification datasets is lower, e.g. the performance of both CoOp and \ourmeos without class-specific context is lower than the linear probing on FGVCAircraft. 
All these performance comparisons can serve as experimental evidence to demonstrate that multiple local prompts and the OT distance facilitate the prompt learning of vision-language models. The detailed accuracy can be found in the appendix.

\begin{wraptable}{r}{7.5cm}
\vspace{-0.2cm}
\caption{Comparisons on robustness to domain shift.}
\vspace{-0.4cm}
\footnotesize
\centering
\newcolumntype{g}{>{\columncolor{Gray}}r}
\renewcommand\tabcolsep{1pt}
\label{tab:robustness}
    \begin{tabular}{lccccc}
    \toprule
    \multirow{2}{*}{\centering Method}
    & Source & \multicolumn{4}{c}{Target} \\ \cmidrule(lr){2-2} \cmidrule(lr){3-6}
    & ImageNet & -V2 & -Sketch & -A & -R \\
    \midrule
    CLIP + CoOp & 61.91 & 54.26 & 32.47 & 21.78 & 54.21 \\
    CLIP + \ourmeos ($N\!=\!4$) & \textbf{63.01} & \textbf{55.11} & \textbf{33.00} & \textbf{21.86} & \textbf{55.61} \\
    \bottomrule
    \end{tabular}
\vspace{-0.5cm}
\end{wraptable} 




\textbf{Domain generalization}
The robustness also plays a critical role in model applications since the real-world environment may have large domain shifts with the training data. 
Therefore, we conducted a robustness evaluation to investigate the transferability of models learned by \ourmeos. Table~\ref{tab:robustness} summarizes the results of our \ourmeos method and CoOp on four ImageNet-based robustness evaluation datasets. For both methods, we trained the models on ImageNet with 16 shots per class.  For \ourmeos, we set the number of prompts as $N=4$. 
We can observe that \ourmeos outperforms CoOp consistently on both source and target domains. These experimental results demonstrate that the performance improvement of our learning multiple prompts doesn't rely on single-domain overfitting.

\vspace{-5pt}
\subsection{Ablation Studies and More Analysis}
\label{sec:analysis}
\vspace{-5pt}

\begin{table*}[t]
\centering
\vspace{-5pt}
 \caption{\textbf{Ablation studies on few-shot recognition}. \ourmeos: our defined model with $N=4$. 
 CoOp: the baseline method.
 G: respectively matching the global visual feature and multiple textual prompts
 V: applying a constraint to add the variance of prompts.
 E: using different initializations as the ensemble:
 M: using the visual feature map instead of the global visual feature. {More details of different variants can be found in Section~\ref{app:ablation} in the appendix. }}
 \label{tab:ablation}  \vspace{-8pt}
 \centering
 \small \renewcommand{\arraystretch}{1.0}
 \setlength{\tabcolsep}{3pt}
\begin{tabular} {l>{\columncolor{white}[1pt][\tabcolsep]}l>{\columncolor{white}[1pt][\tabcolsep]}ccccc}   
\toprule
Dataset                  & Settings  & 1 shot  & 2 shots & 4 shots & 8 shots  & 16 shots\\ \midrule
\multirow{6}{*}{\centering Caltech101} 
    &  \ourmeos   & \bm{$89.83\pm 0.33$}  & \bm{$90.67\pm 0.21$}  & \bm{$90.80\pm 0.20$}  & \bm{$91.54\pm 0.33$}  & \bm{$92.24\pm 0.38$}  \\
    &  CoOp   & $87.51\pm 1.02$  & $87.84\pm 1.10$  & $89.52\pm 0.80$  & $90.28\pm 0.42$  & $91.99\pm 0.31$  \\
    &  G     & $88.13\pm 0.36$  & $86.98\pm 1.25$  & $88.45\pm 0.79$  & $90.16\pm 0.22$  & $90.72\pm 0.18$  \\
    &  G+V    & $88.28\pm 0.43$  & $87.72\pm 1.25$  & $88.45\pm 0.30$  & $89.82\pm 0.20$  & $92.00\pm 0.13$  \\
    &  G+E    & $88.91\pm 0.38$  & $90.01\pm 0.22$  & $90.41\pm 0.20$  & $90.60\pm 0.10$  & $91.74\pm 0.42$  \\
    &  M      & $69.78\pm 1.75$  & $71.57\pm 1.59$  & $77.18\pm 2.16$  & $81.77\pm 0.47$  & $86.21\pm 0.20$  \\
    &  M+V    & $66.11\pm 8.29$  & $71.45\pm 3.98$  & $79.30\pm 3.96$  & $86.96\pm 0.78$  & $89.80\pm 0.17$  \\ 
                                    \midrule
\multirow{6}{*}{\centering DTD} 
    &  \ourmeos   & \bm{$46.55\pm 2.62$}  & \bm{$51.24\pm 1.95$}  & \bm{$56.03\pm 0.43$}  & \bm{$61.70\pm 0.35$}  & \bm{$65.60\pm 0.82$}  \\
    &  CoOp   & $43.62\pm 1.96$  & $45.35\pm 0.31$  & $53.94\pm 1.37$  & $59.69\pm 0.13$  & $62.51\pm 0.25$  \\
    &  G      & $45.12\pm 1.69$  & $48.39\pm 2.08$  & $54.75\pm 0.48$  & $60.15\pm 0.70$  & $63.59\pm 0.76$  \\
    &  G+V    & $45.90\pm 2.00$  & $48.50\pm 0.99$  & $53.96\pm 0.48$  & $59.69\pm 1.01$  & $63.51\pm 0.66$  \\
    &  G+E    & $46.39\pm 1.00$  & $49.31\pm 0.56$  & $52.99\pm 0.60$  & $60.44\pm 1.64$  & $63.97\pm 0.48$  \\
    &  M      & $13.18\pm 4.57$  & $12.25\pm 3.86$  & $13.00\pm 4.73$  & $20.76\pm 5.42$  & $26.99\pm 1.98$  \\
    &  M+V    & $12.61\pm 5.93$  & $15.11\pm 1.81$  & $20.35\pm 1.33$  & $44.13\pm 2.39$  & $56.85\pm 0.54$  \\ 
\midrule
    \multirow{6}{*}{\centering FOOD101} 
    &  \ourmeos   & \bm{$77.74\pm 0.47$}  &\bm{$77.70\pm 0.02$}  & \bm{$77.21\pm 0.43$}  & \bm{$75.31\pm 0.30$}  & \bm{$77.09\pm 0.18$}  \\
    &  CoOp   & $74.25\pm 1.52$  & $72.61\pm 1.33$  & $73.49\pm 2.03$  & $71.58\pm 0.79$  & $74.48\pm 0.15$  \\
    &  G      & $74.63\pm 0.11$  & $70.15\pm 0.49$  & $70.41\pm 0.46$  & $70.72\pm 0.98$  & $73.68\pm 0.46$  \\
    &  G+V    & $74.83\pm 0.31$  & $70.09\pm 0.85$  & $70.86\pm 0.22$  & $70.80\pm 0.68$  & $73.93\pm 0.35$  \\
    &  G+E    & $75.77\pm 0.62$  & $73.54\pm 0.88$  & $75.82\pm 0.44$  & $72.40\pm 0.50$  & $75.52\pm 0.33$  \\
    &  M      & $52.02\pm 4.86$  & $46.12\pm 1.46$  & $46.86\pm 1.39$  & $53.43\pm 0.88$  & $61.28\pm 0.23$  \\
    &  M+V    & $46.52\pm 1.15$  & $45.95\pm 2.66$  & $53.57\pm 0.83$  & $62.95\pm 0.37$  & $67.63\pm 1.11$  \\ \bottomrule
\end{tabular} \vspace{-8pt}
\end{table*}

\begin{table*}[t]
\centering
 \caption{Parameter analysis for the number of prompts}
 \label{tab:parameter}  \vspace{-5pt}
 \centering
 \small \renewcommand{\arraystretch}{1.0}
 \setlength{\tabcolsep}{3pt}
\begin{tabular} {l>{\columncolor{white}[1pt][\tabcolsep]}l>{\columncolor{white}[1pt][\tabcolsep]}ccccc}   
\toprule
Dataset                  & Settings  & 1 shot  & 2 shots & 4 shots & 8 shots  & 16 shots\\ \midrule
\multirow{4}{*}{\centering Caltech101} 
                                    & N=1 & $88.47\pm 1.15$  & $89.19\pm 0.39$  & $89.70\pm 0.38$  & $90.45\pm 0.24$  & $91.56\pm 0.14$  \\
                                    & N=2 & $88.86\pm 0.51$  & $89.60\pm 0.10$  & $90.60\pm 0.17$  & $91.25\pm 0.65$  & $91.89\pm 0.36$  \\
                                    & N=4 & \bm{$89.83\pm 0.33$}  & \bm{$90.67\pm 0.21$}  & $90.80\pm 0.20$  & \bm{$91.54\pm 0.33$}  & \bm{$92.24\pm 0.38$}  \\
                                    & N=8 & $89.74\pm 0.30$  & $90.18\pm 0.46$  &\bm{$91.02\pm 0.18$}  & $91.28\pm 0.28$  & $92.04\pm 0.29$  \\  \midrule
\multirow{4}{*}{\centering DTD} 
                                    & N=1 & $43.91\pm 0.65$  & $48.21\pm 2.20$  & $53.69\pm 1.10$  & $58.90\pm 0.19$  & $62.85\pm 0.74$  \\
                                    & N=2 & $45.59\pm 2.46$  & $48.06\pm 1.92$  & $55.58\pm 1.71$  & $61.56\pm 0.17$  & $64.60\pm 0.92$  \\
                                    & N=4 & $46.55\pm 2.62$  & $51.24\pm 1.95$  & \bm{$56.03\pm 0.43$}  & $61.70\pm 0.35$  & \bm{$65.60\pm 0.82$}  \\
                                    & N=8 & \bm{$46.89\pm 1.94$}  & \bm{$51.87\pm 2.06$}  & $54.45\pm 0.48$  & \bm{$62.20\pm 0.56$}  & $65.25\pm 0.38$  \\ \midrule
    \multirow{4}{*}{\centering FOOD101} 
                                    & N=1 & $75.96\pm 0.48$  & $76.12\pm 0.59$  & $77.11\pm 0.41$  & $76.56\pm 0.69$  & $77.43\pm 0.80$  \\
                                    & N=2 & $77.12\pm 0.49$  & $76.89\pm 0.23$  & $76.16\pm 0.52$  & $75.23\pm 0.69$  & $76.81\pm 0.50$  \\
                                    & N=4 & $77.74\pm 0.47$  & $77.70\pm 0.02$  & $77.21\pm 0.43$  & $75.31\pm 0.30$  & $77.09\pm 0.18$  \\
                                    & N=8 & \bm{$78.05\pm 0.15$}  & \bm{$78.19\pm 0.07$}  & \bm{$78.12\pm 0.17$}  & \bm{$76.63\pm 0.22$}  & \bm{$77.48\pm 0.12$}  \\  \bottomrule
\end{tabular} \vspace{-15pt}
\end{table*}

In this subsection, we conducted the ablation studies to investigate the effectiveness of different components, in order to answer the following questions.

\textbf{Q: Can we directly learn multiple prompts by matching the prompt ensemble with the global visual feature? } \textbf{A: No}.  
As shown in Table~\ref{tab:ablation}, we report the performance of directly matching the prompt ensemble with the global visual feature (notated as ``G'') on three datasets including Caltech101, DTD, and FOOD101.
The performance improvement of this method over CoOp is limited and far lower than \ourmeos. 
It may be because this ``G'' method is incentivized to learn the indistinguishable prompts, which contradicts our purpose to learn multiple comprehensive prompts.  

\textbf{Q: Can ensemble methods that encourage the variety of prompts work well? 
} \textbf{A: Not really}.  
As shown in Table~\ref{tab:ablation}, we further apply two methods to encourage the 
 variety of prompts and then use the ensemble to match the global feature.
In method ``V'', we add an objective function to add the distance between every two prompts as a regularization term.
In method ``E'', we use predefined different initializations to replace the random initializations, such as "a photo of a", "this is a photo", "this is a", and "one picture of a".
However, ``G+V'' did not achieve consistent improvement over the ``G''.
Despite the clear improvement brought by ``G+E'', our \ourmeos showed 
consistent superiority over ``G+E'', which further demonstrates the effectiveness of the OT distance.

\textbf{Q: Does the improvement mainly come from using all feature maps?}  \textbf{A: No}.
In \ourmeos, we apply all feature maps of the visual encoder branch, where each feature is a local embedding at one spatial position. 
However, we demonstrate that the improvement of \ourmeos does not only rely on using all feature maps. On the contrary, directly using the feature map to replace the global feature causes a large performance drop. 
For example, on all three datasets, directly using the feature map (``M'' or ``M+V'') has around $20\%$ 1 shot accuracy drop over using the global visual feature.
It is not surprising since the original CLIP model is trained by matching the global visual feature and language feature. Without using the OT method, the distance between the feature map and multiple textual prompts degenerates to the mean distance of each feature-prompt pair. 


\begin{table*}[t]
\centering
\vspace{-5pt}
 \caption{{The few-shot accuracies of Tip-adapter-F and our adapter-based \ourmeos on 11 datasets.}}
 \vspace{-5pt}
 \label{tab:adapter_all}  
 \centering
 \small \renewcommand{\arraystretch}{1.0}
 \setlength{\tabcolsep}{3pt}
\begin{tabular} {l>{\columncolor{white}[1pt][\tabcolsep]}l>{\columncolor{white}[1pt][\tabcolsep]}ccccc}   
\toprule
Dataset                  & Methods  & 1 shot  & 2 shots & 4 shots & 8 shots  & 16 shots\\ \midrule
\multirow{2}{*}{\centering Caltech101} 
    &  \cellcolor{Gray} Tip-Adapter-F + \ourmeos   & \cellcolor{Gray}$89.33$  & \cellcolor{Gray}$90.87$  & \cellcolor{Gray}$90.87$  & \cellcolor{Gray}$92.29$  &\cellcolor{Gray} $93.18$  \\
    &  Tip-Adapter-F  & $89.33$  & $89.74$  & $90.56$  & $91.44$  & $92.86$  \\
    \midrule
\multirow{2}{*}{\centering DTD} 
    &  \cellcolor{Gray}Tip-Adapter-F + \ourmeos   & \cellcolor{Gray}$51.12$  & \cellcolor{Gray}$52.42$  & \cellcolor{Gray}$59.81$  & \cellcolor{Gray}$63.71$  & \cellcolor{Gray}$67.79$  \\
    &  Tip-Adapter-F  & $49.65$  & $53.72$  &  $57.39$  & $62.71$  & $66.55$  \\ \midrule
    \multirow{2}{*}{\centering EuroSAT} 
    &   \cellcolor{Gray} Tip-Adapter-F +\ourmeos   & \cellcolor{Gray}$64.37$  & \cellcolor{Gray}$76.53$  &\cellcolor{Gray} $79.51$  & \cellcolor{Gray}$79.17$  & \cellcolor{Gray}$85.75$  \\
    &  Tip-Adapter-F  & $59.53$  & $66.15$  & $74.12$  & $77.93$  & $84.54$  \\
    \midrule
    \multirow{2}{*}{\centering FGVCAircraft} 
    &  \cellcolor{Gray}Tip-Adapter-F +\ourmeos   & \cellcolor{Gray}$19.89$  & \cellcolor{Gray}$22.20$  &\cellcolor{Gray} $26.22$  & \cellcolor{Gray}$30.69$  & \cellcolor{Gray}$36.21$  \\
    &  Tip-Adapter-F   & $20.22$  & $23.19$  & $25.80$  & $30.21$  & $35.55$  \\
    \midrule
    \multirow{2}{*}{\centering Flowers102} 
    &  \cellcolor{Gray} Tip-Adapter-F + \ourmeos   & \cellcolor{Gray}$77.59$  & \cellcolor{Gray}$84.98$  & \cellcolor{Gray}$89.32$  & \cellcolor{Gray}$93.75$  & \cellcolor{Gray}$96.10$  \\
    &  Tip-Adapter-F   & $79.98$  & $82.30$  & $88.83$  & $91.51$  & $94.80$  \\
    \midrule
    \multirow{2}{*}{\centering FOOD101} 
    & \cellcolor{Gray} Tip-Adapter-F + \ourmeos   & \cellcolor{Gray}$78.71$  &  \cellcolor{Gray}$78.52$  &  \cellcolor{Gray}$77.90$  & \cellcolor{Gray}$76.93$  &  \cellcolor{Gray}$78.36$  \\
    &  Tip-Adapter-F   & $77.51$  &  $77.81$  &  $78.24$  &  $78.64$  &  $79.43$  \\
    \midrule
    \multirow{2}{*}{\centering ImageNet} 
    &  \cellcolor{Gray} Tip-Adapter-F + \ourmeos   & \cellcolor{Gray}$62.27$  & \cellcolor{Gray}$64.31$  & \cellcolor{Gray}$63.89$  & \cellcolor{Gray}$65.04$  & \cellcolor{Gray}$66.17$  \\
    &  Tip-Adapter-F & 61.32 & 61.69 & 62.52 & 64.00 & 65.51 \\
    \midrule
    \multirow{2}{*}{\centering OxfordPets} 
    &  \cellcolor{Gray} Tip-Adapter-F + \ourmeos   & \cellcolor{Gray}$87.16$  & \cellcolor{Gray}$87.68$  & \cellcolor{Gray}$88.63$  & \cellcolor{Gray}$89.78$  &\cellcolor{Gray} $87.54$  \\
    &  Tip-Adapter-F   & $87.00$  & $87.03$  & $87.54$  & $88.09$  & $89.70$  \\
    \midrule
    \multirow{2}{*}{\centering StanfordCars} 
    &  \cellcolor{Gray} Tip-Adapter-F + \ourmeos   & \cellcolor{Gray}$59.12$  & \cellcolor{Gray}$62.32$  & \cellcolor{Gray}$67.50$  & \cellcolor{Gray}$70.64$  & \cellcolor{Gray}$76.00$  \\
    &  Tip-Adapter-F   & $58.86$  & $61.50$  & $64.57$  & $69.25$  & $75.74$  \\
    \midrule
    \multirow{2}{*}{\centering SUN397} 
    &  \cellcolor{Gray} Tip-Adapter-F + \ourmeos   & \cellcolor{Gray}$64.26$  & \cellcolor{Gray}$64.91$  & \cellcolor{Gray}$67.29$  & \cellcolor{Gray}$69.87$  &\cellcolor{Gray} $71.64$  \\
    &  Tip-Adapter-F   & $62.50$  & $63.64$  & $66.21$  & $68.87$  & $71.47$  \\
    \midrule
    \multirow{2}{*}{\centering UCF101} 
    &  \cellcolor{Gray} Tip-Adapter-F + \ourmeos   & \cellcolor{Gray}$65.69$  \cellcolor{Gray}& $70.22$  & \cellcolor{Gray}$72.56$  &\cellcolor{Gray} $76.53$  & \cellcolor{Gray}$79.51$  \\
    &  Tip-Adapter-F   & $64.87$  & $66.43$  & $70.55$  & $74.25$  & $78.03$  \\
    \midrule
    \multirow{2}{*}{\centering Average} 
    &  \cellcolor{DarkCyan} Tip-Adapter-F + \ourmeos   &  \cellcolor{DarkCyan}$65.45$  &  \cellcolor{DarkCyan}$68.63$  &  \cellcolor{DarkCyan}$71.23$  &  \cellcolor{DarkCyan}$73.49$  &  \cellcolor{DarkCyan}$76.20$  \\
    &  \cellcolor{LightCyan} Tip-Adapter-F   & \cellcolor{LightCyan}$64.62$  & \cellcolor{LightCyan}$66.65$  & \cellcolor{LightCyan}$69.67$  & \cellcolor{LightCyan}$72.45$  & \cellcolor{LightCyan}$75.83$  \\
   \bottomrule
\end{tabular} 
\vspace{-15pt}
\end{table*}

\textbf{Q: How many prompts are needed? } \textbf{A: 4 prompts are enough}
One important hyper-parameter in \ourmeos is the number of prompts. To analyze the effect of the number of prompts, we conducted the experiments on three datasets with $1,2,4,8$ prompts. The results are summarized in the white part of Table~\ref{tab:parameter}.  
We can observe that the performance obviously increases when adding the number of prompts from 1 to 4. For example, \ourmeos (N=4) respectively obtains $1.36\%$, $2.64\%$, and $1.68\%$ 1-shot accuracy improvement over \ourmeos (N=1) on three datasets. Besides, when we further increase the number of prompts, the improvement is not consistent. To balance the improvement and cost, we set $N=4$ as the default configuration of our \ourmeos model. In the experiments, we tuned this hyper-parameter on the Caltech101 dataset and applied it to other datasets.

\textbf{Q: Can \ourmeos benefit Adapter-based methods? } \textbf{A: Yes}.
Adapter-based methods~\citep{clip-adapter,tip-adapter} is another research direction of the efficient adaptation of pre-trained vision-language models. Different from prompt learning that fixes the model parameters and tunes the language prompt, adapter-based methods~\citep{clip-adapter,tip-adapter} allow for fine-tuning a part of the network or adding an extra model for training. Recently, adapter-based methods also achieved good performance on few-shot visual recognition. Therefore, we would like to explore whether our \ourmeos approach can benefit them, and how.



We apply the Tip-adapter-F~\citep{tip-adapter} as our baseline method, which learns a $Linear(d, N_{cls}\times K_{shots})$ model to describe one image by the similarity with all training samples, where $d$ is the dimension of visual feature, $N_{cls}$ is the number of categories (e.g. 1000 in ImageNet), and $K_{shots}$ is the number of shots. Then, the final similarity consists of the original distance between the visual feature and prompt ensembling and the new distance calculated by the learned feature and one-hot vector of labels (whose dimension is $(N_{cls}\times K_{shots}, N_{cls}) $). Please find details in Tip-adapter-F~\citep{tip-adapter}. 
To introduce \ourmeos to this framework, we first used the feature map to replace the global feature and then learned multiple linear models. As a result, with different local features and different linear models, we can obtain a $M \times N$ distance matrix and apply the Sinkhorn algorithm~\citep{cuturi2013sinkhorn} to calculate the OT distance. Furthermore, we can apply the learned prompts as co-partner of the ensembling prompt to refine the final similarity.
Table~\ref{tab:adapter_all} summarizes the few-shot recognition results of the original Tip-Adapter-F method and our adapter-based \ourmeos methods on all 11 datasets.



\textbf{Q: Can \ourmeos benefit zero-shot learning? } \textbf{A: No}. The detailed analysis and discussions can be found in the appendix. 


\textbf{Q: What is the extra computation time cost of PLOT over CoOp baseline?}
\textbf{A: Around \bm{$10\%$} inference speed and \bm{$5\%$} training time}. 
Please see the detailed analysis in the appendix.

\begin{figure}
  \centering
  \includegraphics[width=1.0\linewidth]{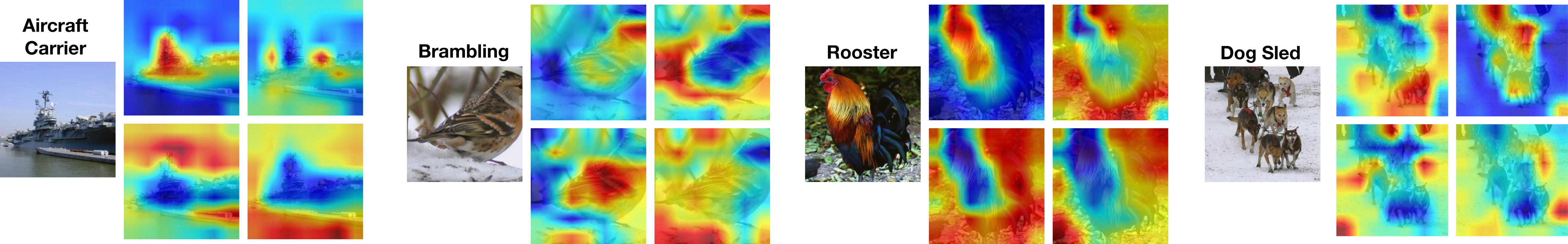}
  \caption{Visualization. We provide the heatmaps of transport plan $\bm{T}$ related to each prompt on 4 categories in ImageNet. Different transport plans focus on different attributes of the object.}
\label{fig: visualization}
\vspace{-0.5cm}
\end{figure}


 
\vspace{-10pt}
\subsection{Visualization}
\vspace{-5pt}
In this subsection, we provide some visualization examples of the transport plans $\bm{T}$ related to different prompts (N=4) in Figure~\ref{fig: visualization}.  A detailed analysis of these visualization examples and further visualization results including the interpretation of the learned prompt, a T-SNE visualization of prompts, and the visualization of the false case can be found in Section~\ref{app: visualization}


\vspace{-10pt}
\section{Conclusion}
\vspace{-5pt}
In this paper, we present a method, named \textbf{\texttt{PLOT}}, to learn multiple comprehensive prompts to describe diverse characteristics of one category. To avoid convergence to one point, we propose to apply the optimal transport to achieve the fine-grained alignment between both vision and language domains. We apply a two-stage optimization strategy where the inner loop fixes the prompts and learns the transport plan to calculate the cross-modality distance, and the outer loop uses this distance to optimize the prompt learner. We build our method on the base of CoOp and achieve significant improvement on the few-shot recognition task in various datasets, which demonstrates the advantage to learn multiple prompts instead of a single one.

\section*{Acknowledgment}
\vspace{-0.25cm}

This project was partially supported by the National Institutes of Health (NIH) under Contract R01HL159805, by the NSF-Convergence Accelerator Track-D award \#2134901, by a grant from Apple Inc., a grant from KDDI Research Inc, and generous gifts from Salesforce Inc., Microsoft Research, and Amazon Research.



\bibliography{plot}

\begin{thebibliography}{69}
\providecommand{\natexlab}[1]{#1}
\providecommand{\url}[1]{\texttt{#1}}
\expandafter\ifx\csname urlstyle\endcsname\relax
  \providecommand{\doi}[1]{doi: #1}\else
  \providecommand{\doi}{doi: \begingroup \urlstyle{rm}\Url}\fi

\bibitem[Arjovsky et~al.(2017)Arjovsky, Chintala, and
  Bottou]{arjovsky2017wasserstein}
Martin Arjovsky, Soumith Chintala, and L{\'e}on Bottou.
\newblock Wasserstein generative adversarial networks.
\newblock In \emph{ICML}, pp.\  214--223, 2017.

\bibitem[Boissard et~al.(2015)Boissard, Le~Gouic, and
  Loubes]{boissard2015distribution}
Emmanuel Boissard, Thibaut Le~Gouic, and Jean-Michel Loubes.
\newblock Distribution’s template estimate with wasserstein metrics.
\newblock \emph{Bernoulli}, 21\penalty0 (2):\penalty0 740--759, 2015.

\bibitem[Bossard et~al.(2014)Bossard, Guillaumin, and Gool]{bossard2014food}
Lukas Bossard, Matthieu Guillaumin, and Luc~Van Gool.
\newblock Food-101--mining discriminative components with random forests.
\newblock In \emph{ECCV}, pp.\  446--461, 2014.

\bibitem[Chen et~al.(2019)Chen, Zhang, Zhang, Tao, Gan, Zhang, Li, Shen, Chen,
  and Carin]{chen2019improving}
Liqun Chen, Yizhe Zhang, Ruiyi Zhang, Chenyang Tao, Zhe Gan, Haichao Zhang, Bai
  Li, Dinghan Shen, Changyou Chen, and Lawrence Carin.
\newblock Improving sequence-to-sequence learning via optimal transport.
\newblock \emph{arXiv preprint arXiv:1901.06283}, 2019.

\bibitem[Cimpoi et~al.(2014)Cimpoi, Maji, Kokkinos, Mohamed, and
  Vedaldi]{cimpoi2014describing}
Mircea Cimpoi, Subhransu Maji, Iasonas Kokkinos, Sammy Mohamed, and Andrea
  Vedaldi.
\newblock Describing textures in the wild.
\newblock In \emph{CVPR}, pp.\  3606--3613, 2014.

\bibitem[Cuturi(2013)]{cuturi2013sinkhorn}
Marco Cuturi.
\newblock Sinkhorn distances: lightspeed computation of optimal transport.
\newblock In \emph{NeurIPS}, volume~2, pp.\ ~4, 2013.

\bibitem[Deng et~al.(2009)Deng, Dong, Socher, Li, Li, and
  Fei-Fei]{deng2009imagenet}
Jia Deng, Wei Dong, Richard Socher, Li-Jia Li, Kai Li, and Li~Fei-Fei.
\newblock Imagenet: A large-scale hierarchical image database.
\newblock In \emph{CVPR}, pp.\  248--255, 2009.

\bibitem[Dou et~al.(2021)Dou, Xu, Gan, Wang, Wang, Wang, Zhu, Liu, Zeng,
  et~al.]{meter}
Zi-Yi Dou, Yichong Xu, Zhe Gan, Jianfeng Wang, Shuohang Wang, Lijuan Wang,
  Chenguang Zhu, Zicheng Liu, Michael Zeng, et~al.
\newblock An empirical study of training end-to-end vision-and-language
  transformers.
\newblock \emph{arXiv preprint arXiv:2111.02387}, 2021.

\bibitem[Fei-Fei et~al.(2004)Fei-Fei, Fergus, and Perona]{fei2004learning}
Li~Fei-Fei, Rob Fergus, and Pietro Perona.
\newblock Learning generative visual models from few training examples: An
  incremental bayesian approach tested on 101 object categories.
\newblock In \emph{CVPRW}, pp.\  178--178, 2004.

\bibitem[Gao et~al.(2021)Gao, Geng, Zhang, Ma, Fang, Zhang, Li, and
  Qiao]{clip-adapter}
Peng Gao, Shijie Geng, Renrui Zhang, Teli Ma, Rongyao Fang, Yongfeng Zhang,
  Hongsheng Li, and Yu~Qiao.
\newblock Clip-adapter: Better vision-language models with feature adapters.
\newblock \emph{arXiv preprint arXiv:2110.04544}, 2021.

\bibitem[He et~al.(2016)He, Zhang, Ren, and Sun]{he2016deep}
Kaiming He, Xiangyu Zhang, Shaoqing Ren, and Jian Sun.
\newblock Deep residual learning for image recognition.
\newblock In \emph{CVPR}, pp.\  770--778, 2016.

\bibitem[Helber et~al.(2019)Helber, Bischke, Dengel, and
  Borth]{helber2019eurosat}
Patrick Helber, Benjamin Bischke, Andreas Dengel, and Damian Borth.
\newblock Eurosat: A novel dataset and deep learning benchmark for land use and
  land cover classification.
\newblock \emph{IEEE Journal of Selected Topics in Applied Earth Observations
  and Remote Sensing}, 12\penalty0 (7):\penalty0 2217--2226, 2019.

\bibitem[Hendrycks et~al.(2019)Hendrycks, Zhao, Basart, Steinhardt, and
  Song]{imagenet-a}
Dan Hendrycks, Kevin Zhao, Steven Basart, Jacob Steinhardt, and Dawn Song.
\newblock Natural adversarial examples.
\newblock \emph{arXiv preprint arXiv:1907.07174}, 2019.

\bibitem[Hendrycks et~al.(2020)Hendrycks, Basart, Mu, Kadavath, Wang, Dorundo,
  Desai, Zhu, Parajuli, Guo, et~al.]{imagenet-r}
Dan Hendrycks, Steven Basart, Norman Mu, Saurav Kadavath, Frank Wang, Evan
  Dorundo, Rahul Desai, Tyler Zhu, Samyak Parajuli, Mike Guo, et~al.
\newblock The many faces of robustness: A critical analysis of
  out-of-distribution generalization.
\newblock \emph{arXiv preprint arXiv:2006.16241}, 2020.

\bibitem[Hong et~al.(2021)Hong, Wu, Qi, Rodriguez-Opazo, and Gould]{vln-bert}
Yicong Hong, Qi~Wu, Yuankai Qi, Cristian Rodriguez-Opazo, and Stephen Gould.
\newblock Vln bert: A recurrent vision-and-language bert for navigation.
\newblock In \emph{CVPR}, pp.\  1643--1653, 2021.

\bibitem[Jain et~al.(2021)Jain, Guo, Srinivasan, Chen, Kudugunta, Jia, Yang,
  and Baldridge]{mural}
Aashi Jain, Mandy Guo, Krishna Srinivasan, Ting Chen, Sneha Kudugunta, Chao
  Jia, Yinfei Yang, and Jason Baldridge.
\newblock Mural: multimodal, multitask retrieval across languages.
\newblock \emph{arXiv preprint arXiv:2109.05125}, 2021.

\bibitem[Jia et~al.(2021)Jia, Yang, Xia, Chen, Parekh, Pham, Le, Sung, Li, and
  Duerig]{align}
Chao Jia, Yinfei Yang, Ye~Xia, Yi-Ting Chen, Zarana Parekh, Hieu Pham, Quoc Le,
  Yun-Hsuan Sung, Zhen Li, and Tom Duerig.
\newblock Scaling up visual and vision-language representation learning with
  noisy text supervision.
\newblock In \emph{ICML}, pp.\  4904--4916, 2021.

\bibitem[Jiang et~al.(2020)Jiang, Xu, Araki, and Neubig]{jiang2020can}
Zhengbao Jiang, Frank~F Xu, Jun Araki, and Graham Neubig.
\newblock How can we know what language models know?
\newblock \emph{TACL}, 8:\penalty0 423--438, 2020.

\bibitem[Kamath et~al.(2021)Kamath, Singh, LeCun, Synnaeve, Misra, and
  Carion]{kamath2021mdetr}
Aishwarya Kamath, Mannat Singh, Yann LeCun, Gabriel Synnaeve, Ishan Misra, and
  Nicolas Carion.
\newblock Mdetr-modulated detection for end-to-end multi-modal understanding.
\newblock In \emph{ICCV}, pp.\  1780--1790, 2021.

\bibitem[Kim et~al.(2021)Kim, Son, and Kim]{vilt}
Wonjae Kim, Bokyung Son, and Ildoo Kim.
\newblock Vilt: Vision-and-language transformer without convolution or region
  supervision.
\newblock In \emph{ICML}, pp.\  5583--5594, 2021.

\bibitem[Krause et~al.(2013)Krause, Stark, Deng, and Fei-Fei]{cars}
Jonathan Krause, Michael Stark, Jia Deng, and Li~Fei-Fei.
\newblock 3d object representations for fine-grained categorization.
\newblock In \emph{ICCVW}, pp.\  554--561, 2013.

\bibitem[Laclau et~al.(2017)Laclau, Redko, Matei, Bennani, and
  Brault]{laclau2017co}
Charlotte Laclau, Ievgen Redko, Basarab Matei, Younes Bennani, and Vincent
  Brault.
\newblock Co-clustering through optimal transport.
\newblock In \emph{ICML}, pp.\  1955--1964, 2017.

\bibitem[Li et~al.(2021)Li, Selvaraju, Gotmare, Joty, Xiong, and
  Hoi]{li2021align}
Junnan Li, Ramprasaath Selvaraju, Akhilesh Gotmare, Shafiq Joty, Caiming Xiong,
  and Steven Chu~Hong Hoi.
\newblock Align before fuse: Vision and language representation learning with
  momentum distillation.
\newblock \emph{NeurIPS}, 34, 2021.

\bibitem[Li et~al.(2022)Li, Li, Xiong, and Hoi]{blip}
Junnan Li, Dongxu Li, Caiming Xiong, and Steven Hoi.
\newblock Blip: Bootstrapping language-image pre-training for unified
  vision-language understanding and generation.
\newblock \emph{arXiv preprint arXiv:2201.12086}, 2022.

\bibitem[Li et~al.(2019)Li, Yatskar, Yin, Hsieh, and Chang]{visualbert}
Liunian~Harold Li, Mark Yatskar, Da~Yin, Cho-Jui Hsieh, and Kai-Wei Chang.
\newblock Visualbert: A simple and performant baseline for vision and language.
\newblock \emph{arXiv preprint arXiv:1908.03557}, 2019.

\bibitem[Li \& Liang(2021)Li and Liang]{li2021prefix}
Xiang~Lisa Li and Percy Liang.
\newblock Prefix-tuning: Optimizing continuous prompts for generation.
\newblock \emph{arXiv preprint arXiv:2101.00190}, 2021.

\bibitem[Liu et~al.(2021{\natexlab{a}})Liu, Rao, Lu, Zhou, and
  Hsieh]{liu2021multi}
Benlin Liu, Yongming Rao, Jiwen Lu, Jie Zhou, and Cho-Jui Hsieh.
\newblock Multi-proxy wasserstein classifier for image classification.
\newblock In \emph{AAAI}, volume~35, pp.\  8618--8626, 2021{\natexlab{a}}.

\bibitem[Liu et~al.(2021{\natexlab{b}})Liu, Yuan, Fu, Jiang, Hayashi, and
  Neubig]{liu2021pre}
Pengfei Liu, Weizhe Yuan, Jinlan Fu, Zhengbao Jiang, Hiroaki Hayashi, and
  Graham Neubig.
\newblock Pre-train, prompt, and predict: A systematic survey of prompting
  methods in natural language processing.
\newblock \emph{arXiv preprint arXiv:2107.13586}, 2021{\natexlab{b}}.

\bibitem[Liu et~al.(2021{\natexlab{c}})Liu, Zheng, Du, Ding, Qian, Yang, and
  Tang]{liu2021gpt}
Xiao Liu, Yanan Zheng, Zhengxiao Du, Ming Ding, Yujie Qian, Zhilin Yang, and
  Jie Tang.
\newblock Gpt understands, too.
\newblock \emph{arXiv preprint arXiv:2103.10385}, 2021{\natexlab{c}}.

\bibitem[Lu et~al.(2022)Lu, Liu, Zhang, Liu, and Tian]{pdl}
Yuning Lu, Jianzhuang Liu, Yonggang Zhang, Yajing Liu, and Xinmei Tian.
\newblock Prompt distribution learning.
\newblock In \emph{CVPR}, pp.\  5206--5215, 2022.

\bibitem[Maji et~al.(2013)Maji, Rahtu, Kannala, Blaschko, and
  Vedaldi]{maji2013fine}
Subhransu Maji, Esa Rahtu, Juho Kannala, Matthew Blaschko, and Andrea Vedaldi.
\newblock Fine-grained visual classification of aircraft.
\newblock \emph{arXiv preprint arXiv:1306.5151}, 2013.

\bibitem[Monge(1781)]{monge1781memoire}
Gaspard Monge.
\newblock M{\'e}moire sur la th{\'e}orie des d{\'e}blais et des remblais.
\newblock \emph{Histoire de l'Acad{\'e}mie Royale des Sciences de Paris}, 1781.

\bibitem[Nichol et~al.(2021)Nichol, Dhariwal, Ramesh, Shyam, Mishkin, McGrew,
  Sutskever, and Chen]{nichol2021glide}
Alex Nichol, Prafulla Dhariwal, Aditya Ramesh, Pranav Shyam, Pamela Mishkin,
  Bob McGrew, Ilya Sutskever, and Mark Chen.
\newblock Glide: Towards photorealistic image generation and editing with
  text-guided diffusion models.
\newblock \emph{arXiv preprint arXiv:2112.10741}, 2021.

\bibitem[Nilsback \& Zisserman(2008)Nilsback and
  Zisserman]{nilsback2008automated}
Maria-Elena Nilsback and Andrew Zisserman.
\newblock Automated flower classification over a large number of classes.
\newblock In \emph{2008 Sixth Indian Conference on Computer Vision, Graphics \&
  Image Processing}, pp.\  722--729, 2008.

\bibitem[Parkhi et~al.(2012)Parkhi, Vedaldi, Zisserman, and
  Jawahar]{parkhi2012cats}
Omkar~M Parkhi, Andrea Vedaldi, Andrew Zisserman, and CV~Jawahar.
\newblock Cats and dogs.
\newblock In \emph{CVPR}, pp.\  3498--3505, 2012.

\bibitem[Paszke et~al.(2019)Paszke, Gross, Massa, Lerer, Bradbury, Chanan,
  Killeen, Lin, Gimelshein, Antiga, et~al.]{paszke2019pytorch}
Adam Paszke, Sam Gross, Francisco Massa, Adam Lerer, James Bradbury, Gregory
  Chanan, Trevor Killeen, Zeming Lin, Natalia Gimelshein, Luca Antiga, et~al.
\newblock Pytorch: An imperative style, high-performance deep learning library.
\newblock \emph{NeurIPS}, 2019.

\bibitem[Patashnik et~al.(2021)Patashnik, Wu, Shechtman, Cohen-Or, and
  Lischinski]{patashnik2021styleclip}
Or~Patashnik, Zongze Wu, Eli Shechtman, Daniel Cohen-Or, and Dani Lischinski.
\newblock Styleclip: Text-driven manipulation of stylegan imagery.
\newblock In \emph{ICCV}, pp.\  2085--2094, 2021.

\bibitem[Petroni et~al.(2019)Petroni, Rockt{\"a}schel, Lewis, Bakhtin, Wu,
  Miller, and Riedel]{petroni2019language}
Fabio Petroni, Tim Rockt{\"a}schel, Patrick Lewis, Anton Bakhtin, Yuxiang Wu,
  Alexander~H Miller, and Sebastian Riedel.
\newblock Language models as knowledge bases?
\newblock \emph{arXiv preprint arXiv:1909.01066}, 2019.

\bibitem[Peyre \& Cuturi(2019)Peyre and Cuturi]{COTFNT}
Gabriel Peyre and Marco Cuturi.
\newblock Computational optimal transport.
\newblock \emph{Foundations and Trends in Machine Learning}, 11\penalty0
  (5-6):\penalty0 355--607, 2019.

\bibitem[Poerner et~al.(2019)Poerner, Waltinger, and
  Sch{\"u}tze]{poerner2019bert}
Nina Poerner, Ulli Waltinger, and Hinrich Sch{\"u}tze.
\newblock Bert is not a knowledge base (yet): Factual knowledge vs. name-based
  reasoning in unsupervised qa.
\newblock \emph{arXiv preprint arXiv:1911.03681}, 2019.

\bibitem[Radford et~al.(2019)Radford, Wu, Child, Luan, Amodei, Sutskever,
  et~al.]{radford2019language}
Alec Radford, Jeffrey Wu, Rewon Child, David Luan, Dario Amodei, Ilya
  Sutskever, et~al.
\newblock Language models are unsupervised multitask learners.
\newblock \emph{OpenAI blog}, 1\penalty0 (8):\penalty0 9, 2019.

\bibitem[Radford et~al.(2021)Radford, Kim, Hallacy, Ramesh, Goh, Agarwal,
  Sastry, Askell, Mishkin, Clark, et~al.]{clip}
Alec Radford, Jong~Wook Kim, Chris Hallacy, Aditya Ramesh, Gabriel Goh,
  Sandhini Agarwal, Girish Sastry, Amanda Askell, Pamela Mishkin, Jack Clark,
  et~al.
\newblock Learning transferable visual models from natural language
  supervision.
\newblock \emph{arXiv preprint arXiv:2103.00020}, 2021.

\bibitem[Ramesh et~al.(2022)Ramesh, Dhariwal, Nichol, Chu, and
  Chen]{ramesh2022hierarchical}
Aditya Ramesh, Prafulla Dhariwal, Alex Nichol, Casey Chu, and Mark Chen.
\newblock Hierarchical text-conditional image generation with clip latents.
\newblock \emph{arXiv preprint arXiv:2204.06125}, 2022.

\bibitem[Rao et~al.(2021)Rao, Zhao, Chen, Tang, Zhu, Huang, Zhou, and
  Lu]{rao2021denseclip}
Yongming Rao, Wenliang Zhao, Guangyi Chen, Yansong Tang, Zheng Zhu, Guan Huang,
  Jie Zhou, and Jiwen Lu.
\newblock Denseclip: Language-guided dense prediction with context-aware
  prompting.
\newblock \emph{arXiv preprint arXiv:2112.01518}, 2021.

\bibitem[Recht et~al.(2019)Recht, Roelofs, Schmidt, and Shankar]{ImageNetv2}
Benjamin Recht, Rebecca Roelofs, Ludwig Schmidt, and Vaishaal Shankar.
\newblock Do imagenet classifiers generalize to imagenet?
\newblock \emph{arXiv preprint arXiv:1902.10811}, 2019.

\bibitem[Rubner et~al.(2000)Rubner, Tomasi, and Guibas]{emd}
Yossi Rubner, Carlo Tomasi, and Leonidas~J Guibas.
\newblock The earth mover's distance as a metric for image retrieval.
\newblock \emph{IJCV}, 40\penalty0 (2):\penalty0 99--121, 2000.

\bibitem[Salimans et~al.(2018)Salimans, Zhang, Radford, and
  Metaxas]{salimans2018improving}
Tim Salimans, Han Zhang, Alec Radford, and Dimitris Metaxas.
\newblock Improving gans using optimal transport.
\newblock \emph{ICLR}, 2018.

\bibitem[Shin et~al.(2020)Shin, Razeghi, Logan~IV, Wallace, and
  Singh]{autoprompt}
Taylor Shin, Yasaman Razeghi, Robert~L Logan~IV, Eric Wallace, and Sameer
  Singh.
\newblock Autoprompt: Eliciting knowledge from language models with
  automatically generated prompts.
\newblock \emph{arXiv preprint arXiv:2010.15980}, 2020.

\bibitem[Soomro et~al.(2012)Soomro, Zamir, and Shah]{soomro2012ucf101}
Khurram Soomro, Amir~Roshan Zamir, and Mubarak Shah.
\newblock Ucf101: A dataset of 101 human actions classes from videos in the
  wild.
\newblock \emph{arXiv preprint arXiv:1212.0402}, 2012.

\bibitem[Tevet et~al.(2022)Tevet, Gordon, Hertz, Bermano, and
  Cohen-Or]{tevet2022motionclip}
Guy Tevet, Brian Gordon, Amir Hertz, Amit~H Bermano, and Daniel Cohen-Or.
\newblock Motionclip: Exposing human motion generation to clip space.
\newblock \emph{arXiv preprint arXiv:2203.08063}, 2022.

\bibitem[Thorpe(2019)]{thorpe2019introduction}
Matthew Thorpe.
\newblock Introduction to optimal transport.
\newblock \emph{Lecture Notes}, 2019.

\bibitem[Tsimpoukelli et~al.(2021)Tsimpoukelli, Menick, Cabi, Eslami, Vinyals,
  and Hill]{tsimpoukelli2021multimodal}
Maria Tsimpoukelli, Jacob~L Menick, Serkan Cabi, SM~Eslami, Oriol Vinyals, and
  Felix Hill.
\newblock Multimodal few-shot learning with frozen language models.
\newblock \emph{NeurIPS}, 34:\penalty0 200--212, 2021.

\bibitem[Tu et~al.(2022)Tu, Zhang, Kjellstr{\"o}m, and Zhang]{tu2022optimal}
Ruibo Tu, Kun Zhang, Hedvig Kjellstr{\"o}m, and Cheng Zhang.
\newblock Optimal transport for causal discovery.
\newblock \emph{ICLR}, 2022.

\bibitem[Van~der Maaten \& Hinton(2008)Van~der Maaten and Hinton]{tsne}
Laurens Van~der Maaten and Geoffrey Hinton.
\newblock Visualizing data using t-sne.
\newblock \emph{JMLR}, 9\penalty0 (11), 2008.

\bibitem[Villani(2009)]{villani2009optimal}
C{\'e}dric Villani.
\newblock \emph{Optimal transport: old and new}, volume 338.
\newblock Springer, 2009.

\bibitem[Wang et~al.(2019)Wang, Ge, Lipton, and Xing]{wang2019learning}
Haohan Wang, Songwei Ge, Zachary Lipton, and Eric~P Xing.
\newblock Learning robust global representations by penalizing local predictive
  power.
\newblock \emph{NeurIPS}, 32, 2019.

\bibitem[Wang et~al.(2021{\natexlab{a}})Wang, Xing, and
  Liu]{wang2021actionclip}
Mengmeng Wang, Jiazheng Xing, and Yong Liu.
\newblock Actionclip: A new paradigm for video action recognition.
\newblock \emph{arXiv preprint arXiv:2109.08472}, 2021{\natexlab{a}}.

\bibitem[Wang et~al.(2021{\natexlab{b}})Wang, Bao, Dong, and Wei]{vlmo}
Wenhui Wang, Hangbo Bao, Li~Dong, and Furu Wei.
\newblock Vlmo: Unified vision-language pre-training with
  mixture-of-modality-experts.
\newblock \emph{arXiv preprint arXiv:2111.02358}, 2021{\natexlab{b}}.

\bibitem[Xiao et~al.(2010)Xiao, Hays, Ehinger, Oliva, and
  Torralba]{xiao2010sun}
Jianxiong Xiao, James Hays, Krista~A Ehinger, Aude Oliva, and Antonio Torralba.
\newblock Sun database: Large-scale scene recognition from abbey to zoo.
\newblock In \emph{CVPR}, pp.\  3485--3492, 2010.

\bibitem[Xu et~al.(2019)Xu, Luo, Zha, and Duke]{xu2019gromov}
Hongteng Xu, Dixin Luo, Hongyuan Zha, and Lawrence~Carin Duke.
\newblock Gromov-wasserstein learning for graph matching and node embedding.
\newblock In \emph{ICML}, pp.\  6932--6941, 2019.

\bibitem[Xu et~al.(2020)Xu, Zhou, Gan, Zheng, and Li]{xu2020vocabulary}
Jingjing Xu, Hao Zhou, Chun Gan, Zaixiang Zheng, and Lei Li.
\newblock Vocabulary learning via optimal transport for neural machine
  translation.
\newblock \emph{arXiv preprint arXiv:2012.15671}, 2020.

\bibitem[Zhang et~al.(2020)Zhang, Cai, Lin, and Shen]{zhang2020deepemd}
Chi Zhang, Yujun Cai, Guosheng Lin, and Chunhua Shen.
\newblock Deepemd: Few-shot image classification with differentiable earth
  mover's distance and structured classifiers.
\newblock In \emph{CVPR}, pp.\  12203--12213, 2020.

\bibitem[Zhang et~al.(2021{\natexlab{a}})Zhang, Fang, Gao, Zhang, Li, Dai,
  Qiao, and Li]{tip-adapter}
Renrui Zhang, Rongyao Fang, Peng Gao, Wei Zhang, Kunchang Li, Jifeng Dai,
  Yu~Qiao, and Hongsheng Li.
\newblock Tip-adapter: Training-free clip-adapter for better vision-language
  modeling.
\newblock \emph{arXiv preprint arXiv:2111.03930}, 2021{\natexlab{a}}.

\bibitem[Zhang et~al.(2021{\natexlab{b}})Zhang, Qiu, Zhang, and Zeng]{vtclip}
Renrui Zhang, Longtian Qiu, Wei Zhang, and Ziyao Zeng.
\newblock Vt-clip: Enhancing vision-language models with visual-guided texts.
\newblock \emph{arXiv preprint arXiv:2112.02399}, 2021{\natexlab{b}}.

\bibitem[Zhao et~al.(2021{\natexlab{a}})Zhao, Phung, Huynh, Le, and
  Buntine]{zhao2020neural}
He~Zhao, Dinh Phung, Viet Huynh, Trung Le, and Wray Buntine.
\newblock Neural topic model via optimal transport.
\newblock \emph{ICLR}, 2021{\natexlab{a}}.

\bibitem[Zhao et~al.(2021{\natexlab{b}})Zhao, Rao, Wang, Lu, and
  Zhou]{zhao2021towards}
Wenliang Zhao, Yongming Rao, Ziyi Wang, Jiwen Lu, and Jie Zhou.
\newblock Towards interpretable deep metric learning with structural matching.
\newblock In \emph{ICCV}, pp.\  9887--9896, 2021{\natexlab{b}}.

\bibitem[Zhou et~al.(2021{\natexlab{a}})Zhou, Loy, and Dai]{zhou2021denseclip}
Chong Zhou, Chen~Change Loy, and Bo~Dai.
\newblock Denseclip: Extract free dense labels from clip.
\newblock \emph{arXiv preprint arXiv:2112.01071}, 2021{\natexlab{a}}.

\bibitem[Zhou et~al.(2021{\natexlab{b}})Zhou, Yang, Loy, and Liu]{coop}
Kaiyang Zhou, Jingkang Yang, Chen~Change Loy, and Ziwei Liu.
\newblock Learning to prompt for vision-language models.
\newblock \emph{arXiv preprint arXiv:2109.01134}, 2021{\natexlab{b}}.

\bibitem[Zhou et~al.(2022)Zhou, Yang, Loy, and Liu]{cocoop}
Kaiyang Zhou, Jingkang Yang, Chen~Change Loy, and Ziwei Liu.
\newblock Conditional prompt learning for vision-language models.
\newblock In \emph{CVPR}, 2022.

\end{thebibliography}
\bibliographystyle{iclr2023_conference}

\clearpage
\appendix

  \textit{\large Appendix for}\\ \ \\
      {\large \bf ``\ourtitle''}\
\vspace{.1cm}

\newcommand{\beginsupplement}{%
	\setcounter{table}{0}
	\renewcommand{\thetable}{A\arabic{table}}%
	\setcounter{figure}{0}
	\renewcommand{\thefigure}{A\arabic{figure}}%
	\setcounter{algorithm}{0}
	\renewcommand{\thealgorithm}{A\arabic{algorithm}}%
	\setcounter{section}{0}
	\renewcommand{\thesection}{A\arabic{section}}%
}

\beginsupplement

{\large Appendix organization:}

\DoToC 

\section{Method Details \label{app:ot}}

The Optimal Transport~\citep{monge1781memoire} is initially introduced to find a transportation plan to move simultaneously several items at a minimal cost, such as moving a pile of sand to fill all the holes. Recently, it is widely used for the comparison of distributions.
Mathematically, given two probability density function $U$ and $V$ over space $\mathcal{X} $ and $\mathcal{Y} $, the OT (Wasserstein) distance~\citep{thorpe2019introduction} can be defined as
\begin{equation} \label{eq:wasserstein}
D_{\text{OT}}(U,V) = \underset{\Gamma}{\inf}  \int_{\mathcal{X}\times\mathcal{Y} } \bm{C}(\bm{x},\bm{y}) d\gamma(\bm{x},\bm{y}),
\end{equation}
where $\bm{C}(\bm{x},\bm{y})$ is the cost between two points in the space $\mathcal{X}\times\mathcal{Y} $, and $\Gamma $ denotes the set of transport plans between support points $ \bm{x}$ and $ \bm{y}$ (e.g. $\gamma(\bm{x},\bm{y}) $). 
We can regard two probability density functions $U$ and $V$ as piles and holes and $\bm{C}$ is the cost function of moving a unit of sand.

In our problem of multiple prompts learning, we formulate the sets of visual features and prompt features as two discrete distributions as
\begin{equation} \label{eq:discrete2}
U=\sum_{m=1}^{M}u_m\delta_{\bm{f}_m} \hspace{2em} \text{and} \hspace{2em} V=\sum_{n=1}^{N}v_n\delta_{\bm{g}_n},
\end{equation}
where $\bm{u}$ and $\bm{v}$ are the discrete probability vectors that sum to 1, and $\delta_{\bm{f}}$ is a Dirac delta function placed at support point $\bm{f}$ in the embedding space. Given two support points $\bm{f}_m$ and $\bm{g}_n$, the cost function is written as $\bm{C} (\bm{f}_m,\bm{g}_n )= 1- \text{sim}(\bm{f}_m,\bm{g}_{n})= 1-\frac{\bm{f}_m^\top\bm{g}_{n}}{||\bm{f}_m||\cdot||\bm{g}_{n}||}$. For simply, in this discrete situation, $\bm{C} \in \mathbb{R}^{M\times N}$ is a cost matrix in which each point denotes the cost between $\bm{f}_m$ and $\bm{g}_n$.
Then, the total distance of these two distributions is written as:
\begin{equation} \label{eq:cost2}
<\bm{T},\bm{C}> = \sum_{m=1}^{M}\sum_{n=1}^{N}\bm{T}_{m,n}\bm{C}_{m,n},
\end{equation}
where the $\bm{T} \in \mathbb{R}^{M\times N} $ is a matrix of transport plan, which is learned to minimize the total distance. Each point $\bm{T}_{m,n}$ in $\bm{T} $ is a weight of local cost $\bm{C}_{m,n} $.

The optimization problem of optimal transport is formulated as:
\begin{equation} \label{eq:optimization2}
\begin{aligned}
& d_{\text{OT}}(\bm{u},\bm{v}|\bm{C}) = \underset{\bm{T}}{\text{minimize}}
 <\bm{T},\bm{C}> \\
& \text{subject to}
~~~~~~\bm{T}\bm{1}_N = \bm{u},\; \bm{T}^\top \bm{1}_M = \bm{v},\; \bm{T} \in \R^{M \times N}_+.
\end{aligned}
\end{equation}
These constraints of $\bm{T}$ are used to match its marginal distributions and original discrete distributions in Eq.~\ref{eq:discrete2}. In our framework, we 
treat visual features $\bm{f}_m$ and prompt features $\bm{g}_n$ equally and thus $\bm{u} = \bm{1}_{M\times 1}/M$ and $\bm{v} = \bm{1}_{N\times 1}/N$.

As directly optimizing the above objective is always time-consuming, we apply the Sinkhorn distance~\citep{cuturi2013sinkhorn} to use an entropic constraint for fast optimization.
The optimization problem with a Lagrange multiplier of the entropy
constraint is:
\begin{equation} \label{eq:Sinkhorn-app-1}
\begin{aligned}
& d_{\text{OT},\lambda}(\bm{u},\bm{v}|\bm{C})=\underset{\bm{T}}{\text{minimize}}
 <\bm{T},\bm{C}> - \lambda h(\bm{T})\\
& \text{subject to}
~~~~~~~\bm{T}\bm{1}_N = \bm{u},\; \bm{T}^\top \bm{1}_M = \bm{v},\; \bm{T} \in \R^{M \times N}_+,
\end{aligned}
\end{equation}
where $h(\cdot) $ is entropy and $\lambda \geq 0$ is a hyper-parameter. Then we can have a fast optimization solution with a few iterations as:
\begin{equation} \label{eq:Sinkhorn-app-2}
\bm{T}^*= \text{diag}(\bm{u}^{(t)}) \exp(-\bm{C}/\lambda) \text{diag}(\bm{v}^{(t)}),
\end{equation}
where $t$ denotes iteration and in each iteration 
$\bm{u}^{(t)} =\bm{u}/\left((\exp(-\bm{C}/\lambda)\bm{v}^{(t-1)}\right) $ and  $\bm{v}^{(t)} =\bm{v}/\left((\exp(-\bm{C}/\lambda)^\top\bm{u}^{(t)}\right) $, with the initiation $\bm{v}^{(0)} = \bm{1} $. The detailed algorithms of the training and testing processes are shown in Algorithms~\ref{Algorithm} and ~\ref{Algorithm2}

\renewcommand{\algorithmicrequire}{\textbf{Input:}}
\renewcommand{\algorithmicensure}{\textbf{Output:}}
\renewcommand{\algorithmicrequire}{\textbf{Input:}}
\begin{algorithm}[tb]
\caption{\!\!\! \textbf{:} The training process of Prompt Learning with Optimal Transport}

\label{Algorithm}
    \begin{algorithmic}[1]
     \Require Training few-shot image data: $\mathbf{X}$ = $\{ \bm{x}\}$, 
     pretrained CLIP model $f$ and $g$, 
      number of prompts $N$, entropy parameter $\lambda$, maximum number of iterations in inner and outer loops $T_{in},T_{out}$.
    \Ensure  The parameters of prompts  $\{\bm{ \omega }_{n}|_{n=1}^{N}\}$ 
    \State {Initialize $\{\bm{ \omega }_{n}|_{n=1}^{N}\}$}
    \For {$t_{out}=1,2,\dots,T_{out}$ in the outer loop}
        \State{Obtain a visual feature set $\bm{F} \in \mathbb{R}^{M \times C} $ with the visual encoder $f(x)$; }
        \State{Generate prompt feature set $\bm{G}_k \in \mathbb{R}^{N \times C}$
        of each class with the textual encoder $\{g(t_k^n)\}|_{n=1}^{N}$;}
        \State{Calculate the cost matrix $\bm{C}_k=\bm{1}-\bm{F}^\top\bm{G}_k \in \mathbb{R}^{M \times N} $ of each class}
        \State{Calculate the OT distance with an inner loop:}
        Initialize the $\bm{v}^{(0)} = \bm{1} $, $\delta =0.01$ and $\Delta_{v}= \infty$
        \For{$t_{in}=1,2,\dots,T_{in}$}
        \State{Update $\bm{u}^{(t_{in})} =\bm{u}/((\exp(-\bm{C}/\lambda)\bm{v}^{(t_{in}-1)}) $}
        \State{Update $\bm{v}^{(t_{in})} =\bm{v}/((\exp(-\bm{C}/\lambda)^\top\bm{u}^{(t_{in})}) $}
        \State{Update $\Delta_{v} =\sum |\bm{v}^{(t_{in})}- \bm{v}^{(t_{in}-1)}|/N $}
        \If{$\Delta_{v}< \delta$}
        \State{break}
        \EndIf
        \EndFor
        \State{Obtain optimal transport plan  as $\bm{T}^*_k= \text{diag}(\bm{u}^{(t)}) \exp(-\bm{C}_k/\lambda) \text{diag}(\bm{v}^{(t)}),$}
        \State{Calculate the OT distance $d_{\text{OT}}(k) = <\bm{T}^*_k,\bm{C}_k >$}
        \State{Calculate the classification probability $p_{\text{OT}}(y=k|\bm{x})$ with the OT distance}
        \State{Update the parameters of prompts $\{\bm{ \omega }_{n}|_{n=1}^{N}\}$ with cross-entropy loss $L_{\text{CE}}$ }
    \EndFor
    \State \Return $\{\bm{ \omega }_{n}|_{n=1}^{N}\}$
    \end{algorithmic}
\end{algorithm}

\renewcommand{\algorithmicrequire}{\textbf{Input:}}
\renewcommand{\algorithmicensure}{\textbf{Output:}}
\renewcommand{\algorithmicrequire}{\textbf{Input:}}
\begin{algorithm}[tb]
\caption{\!\!\! \textbf{:} The inference process of Prompt Learning with Optimal Transport}

\label{Algorithm2}
    \begin{algorithmic}[1]
     \Require Testing image data: $\mathbf{X}$ = $\{ \bm{x}\}$, number of prompts $N$, number of classes $K$,
     learned prompts $\{\bm{t}_k^n|_{k=1,n=1}^{K,N}\}$, 
      a frozen pretrained CLIP model including image encoder $f$ and text encoder$g$
    \Ensure  The classification of each image
    \For {$\bm{x}$ in $\mathbf{X}$ }
        \State{Obtain a visual feature set $\bm{F} \in \mathbb{R}^{M \times C} $ with the visual encoder $f(x)$; }
        \State{Generate prompt feature set $\bm{G}_k \in \mathbb{R}^{N \times C}$
        of each class with the textual encoder $\{g(t_k^n)\}|_{n=1}^{N}$;}
        \State{Calculate the cost matrix $\bm{C}_k=\bm{1}-\bm{F}^\top\bm{G}_k \in \mathbb{R}^{M \times N} $ of each class}
        \State{Calculate the OT distance with an inner loop:}
        Initialize the $\bm{v}^{(0)} = \bm{1} $, $\delta =0.01$ and $\Delta_{v}= \infty$
        \For{$t_{in}=1,2,\dots,T_{in}$}
        \State{Update $\bm{u}^{(t_{in})} =\bm{u}/((\exp(-\bm{C}/\lambda)\bm{v}^{(t_{in}-1)}) $}
        \State{Update $\bm{v}^{(t_{in})} =\bm{v}/((\exp(-\bm{C}/\lambda)^\top\bm{u}^{(t_{in})}) $}
        \State{Update $\Delta_{v} =\sum |\bm{v}^{(t_{in})}- \bm{v}^{(t_{in}-1)}|/N $}
        \If{$\Delta_{v}< \delta$}
        \State{break}
        \EndIf
        \EndFor
        \State{Obtain optimal transport plan  as $\bm{T}^*_k= \text{diag}(\bm{u}^{(t)}) \exp(-\bm{C}_k/\lambda) \text{diag}(\bm{v}^{(t)}),$}
        \State{Calculate the OT distance $d_{\text{OT}}(k) = <\bm{T}^*_k,\bm{C}_k >$}
        \State{Calculate the classification probability $p_{\text{OT}}(y=k|\bm{x})$ with the OT distance}
        \State \Return $k^* = \max\limits_{k} p_{\text{OT}}(y=k|\bm{x}) $
    \EndFor
    \end{algorithmic}
\end{algorithm}


\section{Experimental Details}\label{app:implementation}

\subsection{Dataset Details }\label{app:dataset}
The datasets we used in the experiments follow CoOp~\citep{coop}, which include 11 datasets for few-shot visual recognition and 4 ImageNet-based datasets for generalization (robustness) evaluation. The details of each dataset are shown in Table~\ref{tab:dataset}, including the number of classes, the sizes of training and testing sets, and the original tasks.

\subsection{Implementation Details }\label{app: imple}
The original CoOp method has different versions with different class token positions and parameter initialization strategies. As the performance gap among different versions is limited, we directly chose one of them as our baseline, where the token position is ``end'', the parameter initialization strategy is ``random'', and the length of learnable context tokens is set as 16.
Following the widely used setting in~\citep{coop,cocoop,clip-adapter,tip-adapter}, we also chose RN50~\citep{he2016deep} as the backbone network of the visual branch. All the code of our method is based on CoOp, which adopted the SGD optimizer with 0.002 initial learning rate, CosineAnnealingLR schedule, and a warmup trick with 1e-5 learning rate. We also followed the epoch strategy to train more epochs for more shots. For small datasets such as FGVCAircraft, OxfordFlowers, and StanfordCars, the batch size is set as 32, while for the larger dataset such as Imagenet and SUN397, the batch size is set as 128.

We apply $N = 4$ prompts for each category and use $M = 7\times 7$ due to the feature map size.  We set the hyper-parameters in the Sinkhorn distances algorithm~\citep{cuturi2013sinkhorn} as $\lambda=0.1$ for all the datasets.  
We set the maximum iteration number of the inner loop as 100 and will early stop the iteration when the average absolute update value $\Lambda <0.01$. We initialize all values in the vector $v$ and $\mu$ as $1/N$ and $1/M$ respectively. 
All models are conducted on the Pytorch~\citep{paszke2019pytorch} 1.7.1 and trained on 4 NVIDIA A100 GPUs. We repeated the experiments three times with different seeds and reported the average.

\subsection{Few-shot Recognition Accuracy}\label{app: few-shot}
In Section 4.3.1, we provide a line chart to show and compare the performance of \ourmeos and CoOp. Here, we provide detailed performance results on all 11 few-shot recognition datasets in Table~\ref{tab:few-shot}, where we use gray for our method and white for CoOp. To highlight, we respectively use dark cyan and light cyan to represent the performance of \ourmeos and CoOp on the average of all 11 datasets. We repeat all experiments 3 times and report the mean and standard deviation in the table.

\begin{table*}[t]
    \tabstyle{6pt}
    \caption{The detailed statistics of datasets used in experiments. }
    \label{tab:dataset}
    \resizebox{\textwidth}{!}{
    \begin{tabular}{lcccc}
    \toprule
Dataset                  & Classes  & Training size  & Testing size & Task \\ \midrule
Caltech101~\citep{fei2004learning} & 100 & 4,128 & 2,465& Object recognition \\
DTD~\citep{cimpoi2014describing}& 47 & 2,820 & 1,692 &  Texture recognition\\ 
EuroSAT~\citep{helber2019eurosat}& 10 & 13,500 & 8,100 & Satellite image recognition \\ FGVCAircraft~\citep{maji2013fine} & 100 & 3,334 & 3,333 & Fine-grained aircraft recognition\\
Flowers102~\citep{nilsback2008automated} & 102 & 4,093 & 2,463 & Fine-grained flowers recognition \\ Food101~\citep{bossard2014food} & 101 & 50,500& 30,300 & Fine-grained food recognition  \\ ImageNet~\citep{deng2009imagenet} & 1,000 & 1.28M & 50,000 & Object recognition \\ OxfordPets~\citep{parkhi2012cats} & 37  & 2,944 & 3,669 & Fine-grained pets recognition \\ StanfordCars~\citep{cars} & 196 & 6,509 & 8,041 & Fine-grained car recognition \\
SUN397~\citep{xiao2010sun}& 397& 15,880 & 19,850 & Scene recognition\\ 
UCF101~\citep{soomro2012ucf101}& 101 & 7,639 & 3,783 & Action recognition\\
\midrule
ImageNetV2~\citep{ImageNetv2} & 1,000 & - & 10,000 & Robustness of collocation  \\
ImageNet-Sketch~\citep{wang2019learning} & 1000 & - &50,889 & Robustness of sketch domain\\
ImageNet-A~\citep{imagenet-a}& 200 & - &7,500 &Robustness of adversarial attack\\
ImageNet-R~\citep{imagenet-r}& 200 & - &30,000&Robustness of multi-domains\\
    \bottomrule
    \end{tabular}
    }
\end{table*}

\begin{table*}[t]
\centering
 \caption{The few-shot visual recognition accuracy on 11 datasets.}
 \label{tab:few-shot}  
 \centering
 \small \renewcommand{\arraystretch}{1.0}
 \setlength{\tabcolsep}{3pt}
\begin{tabular} {l>{\columncolor{white}[1pt][\tabcolsep]}l>{\columncolor{white}[1pt][\tabcolsep]}ccccc}   
\toprule
Dataset                  & Methods  & 1 shot  & 2 shots & 4 shots & 8 shots  & 16 shots\\ \midrule
\multirow{2}{*}{\centering Caltech101} 
    &  \cellcolor{Gray}\ourmeos   & \cellcolor{Gray}$89.83\pm 0.33$  & \cellcolor{Gray}$90.67\pm 0.21$  & \cellcolor{Gray}$90.80\pm 0.20$  & \cellcolor{Gray}$91.54\pm 0.33$  &\cellcolor{Gray} $92.24\pm 0.38$  \\
    &  CoOp   & $87.51\pm 1.02$  & $87.84\pm 1.10$  & $89.52\pm 0.80$  & $90.28\pm 0.42$  & $91.99\pm 0.31$  \\
    \midrule
\multirow{2}{*}{\centering DTD} 
    &  \cellcolor{Gray}\ourmeos   & \cellcolor{Gray}$46.55\pm 2.62$  & \cellcolor{Gray}$51.24\pm 1.95$  & \cellcolor{Gray}$56.03\pm 0.43$  & \cellcolor{Gray}$61.70\pm 0.35$  & \cellcolor{Gray}$65.60\pm 0.82$  \\
    &  CoOp   & $43.62\pm 1.96$  & $45.35\pm 0.31$  &  $53.94\pm 1.37$  & $59.69\pm 0.13$  & $62.51\pm 0.25$  \\ \midrule
    \multirow{2}{*}{\centering EuroSAT} 
    &   \cellcolor{Gray}\ourmeos   & \cellcolor{Gray}$54.05\pm 5.95$  & \cellcolor{Gray}$64.21\pm 1.90$  &\cellcolor{Gray} $72.36\pm 2.29$  & \cellcolor{Gray}$78.15\pm 2.65$  & \cellcolor{Gray}$82.23\pm 0.91$  \\
    &  CoOp   & $52.12\pm 5.46$  & $59.00\pm 3.48$  & $68.61\pm 3.54$  & $77.08\pm 2.42$  & $83.69\pm 0.47$  \\
    \midrule
    \multirow{2}{*}{\centering FGVCAircraft} 
    &  \cellcolor{Gray}\ourmeos   & \cellcolor{Gray}$17.90\pm 0.09$  & \cellcolor{Gray}$18.94\pm 0.44$  &\cellcolor{Gray} $22.36\pm 0.42$  & \cellcolor{Gray}$26.17\pm 0.29$  & \cellcolor{Gray}$31.49\pm 0.89$  \\
    &  CoOp   & $8.59\pm 5.79$  & $16.52\pm 2.38$  & $20.63\pm 2.46$  & $26.63\pm 0.86$  & $31.43\pm 0.96$  \\
    \midrule
    \multirow{2}{*}{\centering Flowers102} 
    &  \cellcolor{Gray}\ourmeos   & \cellcolor{Gray}$71.72\pm 0.97$  & \cellcolor{Gray}$81.19\pm 0.79$  & \cellcolor{Gray}$87.82\pm 0.20$  & \cellcolor{Gray}$92.43\pm 0.25$  & \cellcolor{Gray}$94.76\pm 0.34$  \\
    &  CoOp   & $67.98\pm 1.98$  & $77.58\pm 1.46$  & $86.10\pm 1.05$  & $91.27\pm 0.83$  & $94.49\pm 0.40$  \\
    \midrule
    \multirow{2}{*}{\centering FOOD101} 
    & \cellcolor{Gray}\ourmeos   & \cellcolor{Gray}$77.74\pm 0.47$  &  \cellcolor{Gray}$77.70\pm 0.02$  &  \cellcolor{Gray}$77.21\pm 0.43$  & \cellcolor{Gray}$75.31\pm 0.30$  &  \cellcolor{Gray}$77.09\pm 0.18$  \\
    &  CoOp   & $74.25\pm 1.52$  &  $72.61\pm 1.33$  &  $73.49\pm 2.03$  &  $71.58\pm 0.79$  &  $74.48\pm 0.15$  \\
    \midrule
    \multirow{2}{*}{\centering ImageNet} 
    &  \cellcolor{Gray}\ourmeos   & \cellcolor{Gray}$59.54\pm 0.16$  & \cellcolor{Gray}$60.64\pm 0.06$  & \cellcolor{Gray}$61.49\pm 0.23$  & \cellcolor{Gray}$61.92\pm 0.09$  & \cellcolor{Gray}$63.01\pm 0.13$  \\
    &  CoOp   & $56.99\pm 1.03$  & $56.40\pm 0.87$  & $58.48\pm 0.47$  & $60.39\pm 0.57$  & $61.91\pm 0.17$  \\
    \midrule
    \multirow{2}{*}{\centering OxfordPets} 
    &  \cellcolor{Gray}\ourmeos   & \cellcolor{Gray}$87.49\pm 0.57$  & \cellcolor{Gray}$86.64\pm 0.63$  & \cellcolor{Gray}$88.63\pm 0.26$  & \cellcolor{Gray}$87.39\pm 0.74$  &\cellcolor{Gray} $87.21\pm 0.40$  \\
    &  CoOp   & $85.99\pm 0.28$  & $82.22\pm 2.15$  & $86.65\pm 0.97$  & $85.36\pm 1.00$  & $87.02\pm 0.89$  \\
    \midrule
    \multirow{2}{*}{\centering StanfordCars} 
    &  \cellcolor{Gray}\ourmeos   & \cellcolor{Gray}$56.60\pm 0.36$  & \cellcolor{Gray}$57.52\pm 0.71$  & \cellcolor{Gray}$63.41\pm 0.29$  & \cellcolor{Gray}$67.03\pm 0.50$  & \cellcolor{Gray}$72.80\pm 0.75$  \\
    &  CoOp   & $55.81\pm 1.67$  & $58.41\pm 0.43$  & $62.74\pm 0.16$  & $67.64\pm 0.06$  & $73.60\pm 0.19$  \\
    \midrule
    \multirow{2}{*}{\centering SUN397} 
    &  \cellcolor{Gray}\ourmeos   & \cellcolor{Gray}$62.47\pm 0.43$  & \cellcolor{Gray}$61.71\pm 0.65$  & \cellcolor{Gray}$65.09\pm 0.43$  & \cellcolor{Gray}$67.48\pm 0.04$  &\cellcolor{Gray} $69.96\pm 0.24$  \\
    &  CoOp   & $60.12\pm 0.82$  & $59.60\pm 0.76$  & $63.24\pm 0.63$  & $65.77\pm 0.02$  & $68.36\pm 0.66$  \\
    \midrule
    \multirow{2}{*}{\centering UCF101} 
    &  \cellcolor{Gray}\ourmeos   & \cellcolor{Gray}$64.53\pm 0.70$  \cellcolor{Gray}& $66.83\pm 0.43$  & \cellcolor{Gray}$69.60\pm 0.67$  &\cellcolor{Gray} $74.45\pm 0.50$  & \cellcolor{Gray}$77.26\pm 0.64$  \\
    &  CoOp   & $62.13\pm 1.14$  & $64.05\pm 0.99$  & $67.79\pm 0.71$  & $72.71\pm 0.50$  & $76.90\pm 0.50$  \\
    \midrule
    \multirow{2}{*}{\centering Average} 
    &  \cellcolor{DarkCyan}\ourmeos   &  \cellcolor{DarkCyan}$62.59\pm 1.13$  &  \cellcolor{DarkCyan}$65.23\pm 0.72$  &  \cellcolor{DarkCyan}$68.60\pm 0.52$  &  \cellcolor{DarkCyan}$71.23\pm 0.51$  &  \cellcolor{DarkCyan}$73.94\pm 0.54$  \\
    &  \cellcolor{LightCyan}CoOp   & \cellcolor{LightCyan}$59.56\pm 2.06$  & \cellcolor{LightCyan}$61.78\pm 1.39$  & \cellcolor{LightCyan}$66.47\pm 1.29$  & \cellcolor{LightCyan}$69.85\pm 0.69$  & \cellcolor{LightCyan}$73.33\pm 0.42$  \\
   \bottomrule
\end{tabular} 
\end{table*}

\subsection{Ablation Studies Details }\label{app:ablation}
{
In this section, we provide more details about the different variants in Table~\ref{tab:ablation}. We compare \ourmeos with the other 6 baseline methods briefly described below:
\begin{itemize}
\setlength{\itemsep}{2pt}
\setlength{\parsep}{2pt}
\setlength{\parskip}{2pt}
    \item CoOp: CoOp is the baseline method that only learns a single prompt and matches this single prompt and the global visual feature. We apply the officially released code to reproduce this method.
    \item ``G'': In this paper, we propose to explore whether we can learn multiple prompts for more comprehensive textual representation and fine-grained visual-textual alignment. ``G'' denotes that we build multiple prompts (similar to our \ourmeos) and learn them by matching them with the single global visual feature. 
    \item ``G+V'': Matching all local prompts to a single visual feature will reduce the diversity of the learned prompts. To improve the variety of learned prompts, ``G+V'' further adds an objective function to increase the distances between every two prompts.
    \item ``G+E'': ``G+E'' is also a method to increase the variety of prompts by separated initializations. It applies predefined different initializations to replace the random initialization, such as "a photo of a", "this is a photo", "this is a", and "one picture of a".
    \item ``M'': One key difference between \ourmeos and CoOp is to utilize the feature map for more fine-grained information. To evaluate whether our improvement mainly comes from using a feature map, we design a method ``M'', which removes the OT distance of \ourmeos and matches local visual features and multiple textual prompts by the average distance of each visual-textual pair. 
    \item ``M+V'': Similar to ``G+V'', we add an objective function to increase the distances between every two prompts to the method ``M'' to increase the variety of prompts.
\end{itemize}
}

\subsection{Base-to-New Results} \label{app:cocoop}
{To investigate the generalization of our method for other baseline prompt-learning-based methods, we apply our \ourmeos to CoCoOp~\cite{cocoop}, by learning multiple textual prompts (e.g. N=4) instead of the single prompt in CoCoOp. We name it \textbf{\texttt{CoPLOT}}. Specially, we learn multiple prompts and use the same meta-network for all local prompts. Then we apply the Optimal Transport to calculate the distance between multiple local prompts and local visual features.  We evaluate both CoCoOp and CoPLOT in the setting of "base-to-new" and implement them using the same RN50 backbone. 
The results on the 11 datasets with 16 shots are provided in Table~\ref{tab:results_generalization}. We observe that PLOT achieves improvement on most datasets and on average, which demonstrates that it can be applied to different prompt-learning-based methods. For example, on average,\ourmeos achieves almost $3\% $ improvement on the "new" side without the reduction of "base" performance. 
It suggests that these two methods are complementary: CoCoOp proposes a conditional formulation that uses each image feature as the context condition to refine the single prompt, while PLOT aims to learn multiple prompts.
}



\begin{table*}[t]
    \tabstyle{9pt}
    \centering
    \caption{{\textbf{Comparison of CoCoOp~\cite{cocoop} and CoPLOT(ours) in the base-to-new generalization setting}. All methods are implemented with RN50 backbone and evaluated with 16 shots. We report the performance of the base classes, new classes, and the mean of them.  We show that PLOT can be applied to CoCoOp~\cite{cocoop} and achieve improvement. }}
    \label{tab:results_generalization}
    \begin{subtable}[t]{0.3\textwidth}
    \centering
    \setlength{\tabcolsep}{3pt}
    \caption{\textbf{Average }.}
    \begin{tabular}{l cc|c}
    \toprule
    & Base & New & H \\
    \midrule
    CoCoOp & 75.7 & 64.6 & 70.2 \\
    CoPLOT & 75.9 & 67.6 & 71.8 \\
    \bottomrule
    \end{tabular}
    \end{subtable}
    \vspace{1em}
    \begin{subtable}[t]{.3\textwidth}
    \centering
    \setlength{\tabcolsep}{3pt}
    \caption{ImageNet.}
    \begin{tabular}{l cc|c}
    \toprule
    & Base & New & H \\
    \midrule
    CoCoOp & 68.3 & 63.1 & 65.7 \\
    CoPLOT & 68.2 & 63.1 & 65.7 \\
    \bottomrule
    \end{tabular}
    \end{subtable}
    ~
    \begin{subtable}[t]{.3\textwidth}
    \centering
    \setlength{\tabcolsep}{3pt}
    \caption{Caltech101.}
    \begin{tabular}{l cc|c}
    \toprule
    & Base & New & H \\
    \midrule
    CoCoOp & 95.0 & 90.0 & 92.5\\
    CoPLOT & 95.4 & 90.9 & 93.2 \\
    \bottomrule
    \end{tabular}
    \end{subtable}
    ~
    \begin{subtable}[t]{.3\textwidth}
    \centering
    \setlength{\tabcolsep}{3pt}
    \caption{OxfordPets.}
    \begin{tabular}{l cc|c}
    \toprule
    & Base & New & H \\
    \midrule
    CoCoOp & 92.3 & 94.6 & 93.5 \\
    CoPLOT & 92.1 & 95.9 & 94 \\
    \bottomrule
    \end{tabular}
    \end{subtable}
    \vspace{1em}
    \begin{subtable}[t]{.3\textwidth}
    \centering
    \setlength{\tabcolsep}{3pt}
    \caption{StanfordCars.}
    \begin{tabular}{l cc|c}
    \toprule
    & Base & New & H \\
    \midrule
     CoCoOp & 61.8 & 65.3 & 63.6 \\
     CoPLOT & 63.2 & 66.5& 64.9 \\
    \bottomrule
    \end{tabular}
    \end{subtable}
    ~
    \begin{subtable}[t]{.3\textwidth}
    \centering
    \setlength{\tabcolsep}{3pt}
    \caption{Flowers102.}
    \begin{tabular}{l cc|c}
    \toprule
    & Base & New & H \\
    \midrule
     CoCoOp & 91.2 & 67.5 & 79.4 \\
    CoPLOT & 89.6 & 69.2 & 79.4 \\
    \bottomrule
    \end{tabular}
    \end{subtable}
    ~
    \begin{subtable}[t]{.3\textwidth}
    \centering
    \setlength{\tabcolsep}{3pt}
    \caption{Food101.}
    \begin{tabular}{l cc|c}
    \toprule
    & Base & New & H \\
    \midrule
     CoCoOp & 85.0 & 86 & 85.5 \\
    CoPLOT & 85.0 & 85.2 & 85.1 \\
    \bottomrule
    \end{tabular}
    \end{subtable}
    \vspace{1em}
    \begin{subtable}[t]{.3\textwidth}
    \centering
    \setlength{\tabcolsep}{3pt}
    \caption{FGVCAircraft.}
    \begin{tabular}{l cc|c}
    \toprule
    & Base & New & H \\
    \midrule
     CoCoOp &25.5 & 25.7 & 25.6 \\
    CoPLOT & 25.6 & 26.6 & 26.1 \\
    \bottomrule
    \end{tabular}
    \end{subtable}
    ~
    \begin{subtable}[t]{.3\textwidth}
    \centering
    \setlength{\tabcolsep}{3pt}
    \caption{SUN397.}
    \begin{tabular}{l cc|c}
    \toprule
    & Base & New & H \\
    \midrule
     CoCoOp & 75.1 & 73.6 & 74.4 \\
    CoPLOT & 75.2 & 73.2 & - \\
    \bottomrule
    \end{tabular}
    \end{subtable}
    ~
    \begin{subtable}[t]{.3\textwidth}
    \centering
    \setlength{\tabcolsep}{3pt}
    \caption{DTD.}
    \begin{tabular}{l cc|c}
    \toprule
    & Base & New & H \\
    \midrule
    CoCoOp & 73.1 & 50.0 & 61.6 \\
    CoPLOT & 72.6 & 51.4 & 62.0\\
    \bottomrule
    \end{tabular}
    \end{subtable}
    ~
    \begin{subtable}[t]{.3\textwidth}
    \centering
    \setlength{\tabcolsep}{3pt}
    \caption{EuroSAT.}
    \begin{tabular}{l cc|c}
    \toprule
    & Base & New & H \\
    \midrule
     CoCoOp & 88.9 & 33.5 & 61.2 \\
    CoPLOT & 91.0 & 55.3 & 73.2\\
    \bottomrule
    \end{tabular}
    \end{subtable}
    ~
    \begin{subtable}[t]{.3\textwidth}
    \centering
    \setlength{\tabcolsep}{3pt}
    \caption{UCF101.}
    \begin{tabular}{l cc|c}
    \toprule
    & Base & New & H \\
    \midrule
    CoCoOp & 76.5 & 61.6 & 69.1 \\
    CoPLOT & 77.4 & 66.2 & 71.8\\
    \bottomrule
    \end{tabular}
    \end{subtable}
\end{table*}

\subsection{Zero-shot Setting Analysis}\label{app:zero}
\ourmeos can not benefit in the setting of zero-shot learning. Below we provide some experimental details and corresponding analysis. 
CLIP shows that manually designing the prompts can still achieve good performance. We obtain 7 prompts by prompt engineering on the ImageNet dataset and can further ensemble them to obtain \bm{$60.38\%$} top 1 accuracy. In this section, we replace the cosine distance between the global visual feature and prompt ensemble with the OT distance between the feature map and all 7 prompts. However, without any learning, the OT distance only obtains \bm{$58.78\%$} accuracy. It is a limitation of the \ourmeos to still need few-shot data for optimization, which cannot be directly applied in the zero-shot setting. We argue there are two reasons why the OT distance does not work without learning: 1) prompt engineering selects prompts based on the global feature and cosine distance, instead of OT distance with feature map; 2) all these selected prompts are close to the global feature and lack the complementarity.

\subsection{Computation Cost Evaluation }\label{app:cost}

As shown in Table~\ref{tab:time}, we provide the comparison of the training time and inference seed of the baseline method CoOp~\citep{coop} and our \ourmeos with the different number of prompts. We report the one-epoch time training on the 1-shot setting of the Food101~\citep{bossard2014food} dataset and the number of images processed by the model in 1 second. Taking $N=4$ as an example, \ourmeos only reduces the $9.2\%$ inference speed and requires an extra $4.9\%$ training time, which is acceptable given the performance improvement.


\begin{table*}[t]
    \tabstyle{9pt}
    \caption{The training and inference time comparison.}
    \label{tab:time}
	 \setlength{\tabcolsep}{4pt}
	\begin{tabular}{lccccc}
	 \toprule
Settings& CoOp  & \ourmeos(N=1)  & \ourmeos(N=2) & \ourmeos(N=4) & \ourmeos(N=8) \\ \midrule
Training Time (s) & 1.127 & 1.135 & 1.148 & 1.182 & 1.267 \\
Inference Time (images/s) & 719.1 & 714.4 & 690.7 & 653.0 & 519.8 \\
    \bottomrule
    \end{tabular}
\end{table*}

\begin{table*}[t]
    \tabstyle{9pt}
    \caption{The nearest words for 16 context vectors of all $N=4$ prompts learned by \ourmeos. N/A means non-Latin characters. }
    \label{tab:prompts}
    \begin{tabular}{lcccc}
    \toprule
Number                  & Prompt 1  & Prompt 2  & Prompt 3 & Prompt 4\\ \midrule
1& ag & pa & trying & gaz\\ 
2& flint & as & field & \textbf{white} \\ 
3 & leaving & wit & N/A & t\\
4 & sot & l & icons & ario \\
5 & tint & N/A& eclub & safe  \\ 
6& tar & yl &indiffe&class\\
7 & attn & N/A & ts & represented\\
8 & 2  & job & cold & attend \\
9 & rollingstones & built & yeah & vie \\
10& N/A& brought & band  &recognized\\ 
11& N/A & or& love & old\\
12 & bel & j & late & stel  \\
13 & \textbf{head} & ag &industry & awhile\\
14& artifact &bad &N/A &ded\\
15& an &chie &across &these\\
16& 5 & in &actual &visiting\\
    \bottomrule
    \end{tabular} 
\end{table*}

\section{Visualization} \label{app: visualization}

\subsection{More Analysis on Visualization}  \label{app: analysis_v}
In this section, we provide some visualization examples of the transport plans $\bm{T}$ related to different prompts (N=4).  
We translate each transport plan into colorful heatmaps and resize them to their original size and combine them with the raw image. 
As shown in Figure~\ref{fig: visualization}, we provide the heatmaps of 4 categories in ImageNet. We observe that different transport plans highlight different regions of the image, which demonstrates that the learned multiple prompts are complementary. For the class ``Brambling'', the prompts respectively focus on the head, tail, wing, and environment. 
For ``Dog Sled'', the prompts are related to dogs, the sled, some ties, and the snow environment.  


\subsection{Visualization of Failure Cases}  \label{app: failure}
\begin{figure}[t]
    \centering
    \includegraphics[width=\linewidth]{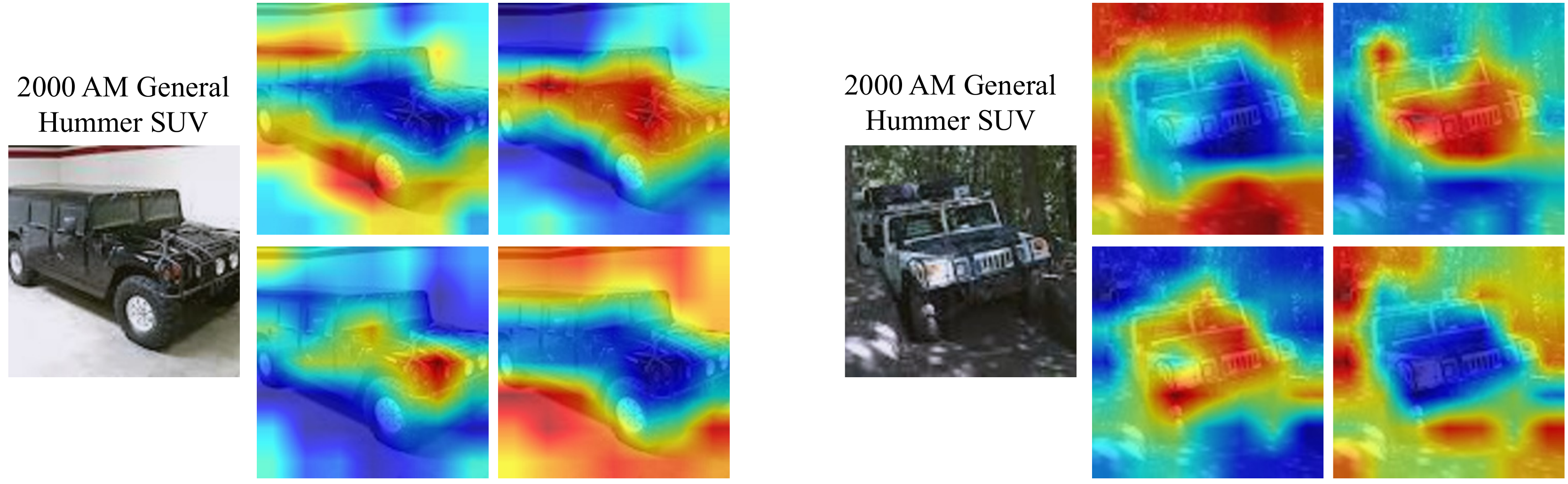}
    \caption{Failure Visualization. We provide the heatmaps of transport plan T related to each prompt on 2 failure examples in the StanfordCars dataset.}
    \label{fig:fail}
\end{figure}

To better understand the method and further discover the reason for the failure cases. we visualize the attention maps of some failure cases. As shown in Figure ~\ref{fig:fail}, we showed two failure examples with class "2000 AM General Hummer" in the StanfordCars dataset. 
During the training, we set the number of prompts as 4, but in these visualization results, we found that some of the learned prompts remarkably coincide with each other. These prompts can be roughly divided into two classes: Foreground and Background. For example, in both images, 
prompts 2 (right top) and 3 (left down) focus on the foreground car, while the others focus on the background. It demonstrates that not all classes have multiple complementary attributes, which motivates us to go further to learn the dynamic local prompts numbers to reduce the computational load in the future.


\subsection{Interpretation of Text Prompts}  \label{app: prompts_v}
The learned prompts are difficult to be understood by humans since the parameters are optimized in the continuous space~\citep{coop}. CoOp proposes to use the word which is nearest to learned prompts in the embedding space to visualize the prompts. Following this manner, we show the nearest words of our learned prompts in Table~\ref{tab:prompts}.  Similar to CoOp, most words can not be directly understood by human logic.  However, we still find the relations between the learned prompts and the corresponding optimal transport plan. As shown in Figure 4 in the main paper, we can observe that the optimal transport plan for Prompt 1 always focuses on the ``head'', such as the head of ``brambling'', the head of ``rooster'', and even the head of ``aircraft carrier''.  It is because the word ``head'' is in Prompt 1.  Similarly, we can find that Prompt 4 prefers the white part of images, such as the white environment in the image of ``brambling'' and the snow in the image of ``dog sled''.  It demonstrates that the learned multiple prompts focus on different characteristics of categories.  

\subsection{T-SNE of Prompts}  \label{app: prompts_v}
\begin{figure}[t]
    \centering
    \includegraphics[width=0.6\linewidth]{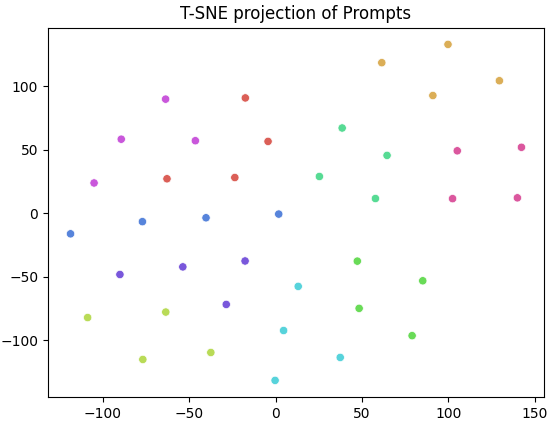}
    \caption{{T-SNE Visualization of 10 classes with different prompts. We apply the T-SNE for the embeddings of 10 randomly selected classes in ImageNet with different prompts. Different color denotes different classes. We observe that the textual embeddings cluster well.}}
    \label{fig: tsne}
\end{figure}

{To better understand the learned prompts, we provide a visualization with T-SNE~\cite{tsne} for the learned textual prompts. Specifically, we randomly select 10 classes from ImageNet and generate the textual embedding with our learned prompts. Then, we obtain $4 \times 10$ embeddings with dimension $d= 1024$. Then we apply the T-SNE to reduce the dimension and visualize the embeddings. As shown in Figure~\ref{fig: tsne}, the textual embeddings of the same class with different prompts are clustered well.  Besides, despite being well clustered, we found that the textual embeddings also have intra-diversities. }


\end{document}